\documentclass[runningheads]{llncs}

\usepackage{multirow}
\usepackage[mobile]{eccv}
\usepackage[pagebackref,breaklinks,colorlinks]{hyperref}
\usepackage{csquotes}

\hypersetup{colorlinks,linkcolor={blue},citecolor={blue},urlcolor={blue}}

\usepackage{eccvabbrv}
\usepackage{epsfig}
\usepackage{svg}
\usepackage{wrapfig}

\usepackage{graphicx}
\usepackage{booktabs}

\usepackage[accsupp]{axessibility}  

\usepackage{pifont} 

\usepackage{hyperref}

\usepackage{tikz}

\begin{document}

\title{Towards Multimodal In-Context Learning for Vision \& Language Models} 


\author{\textbf{Sivan Doveh}\inst{1,2}\quad
\textbf{Shaked Perek}\inst{1}\quad
\textbf{M. Jehanzeb Mirza}\inst{3}\quad
\textbf{Wei Lin}\inst{5}\quad \\
\textbf{Amit Alfassy}\inst{1}\quad
\textbf{Assaf Arbelle} \inst{1}\quad
\textbf{Shimon Ullman} \inst{2}\quad
\textbf{Leonid Karlinsky}\inst{4}
{\institute{$^1$IBM Research \quad $^2$Weizmann Institute of Science \\\quad $^3$ICG,TU Graz \quad $^4$MIT-IBM Watson AI Lab \quad $^5$ELLIS Unit, LIT AI Lab, Institute for Machine Learning, JKU Linz, Austria\\}}
}
\authorrunning{S. Doveh et al.}

\maketitle
\newcommand{\method}{MPVR\xspace}
\newcommand{\ClipEnc}{\theta}
\newcommand{\vis}{\theta_i}  
\newcommand{\txt}{\theta_t}  
\newcommand{\visEnc}{\vis} 
\newcommand{\txtEnc}{\txt} 

\newcommand{\txtEmb}{t_c} 
\newcommand{\txtEmbNew}{t_{\hat{c}}} 

\newcommand{\imgEmb}{\vis} 

\newcommand{\cls}{c} 
\newcommand{\ClassNames}{C} 

\newcommand{\img}{x} 

\newcommand{\prompt}{p} 

\newcommand{\simil}{\cos} 

\newcommand{\likel}{l} 

\newcommand{\LSCE}{\mathcal{L}_\textsc{SCE}}

\newcommand\blue[1]{\textcolor{blue}{#1}}

\newif\ifdraft
\draftfalse
\drafttrue 
\ifdraft
 \newcommand{\MK}[1]{{\color{magenta}{\bf MK: #1}}}
  \newcommand{\JM}[1]{{\color{blue}{\bf JM: #1}}}
\newcommand{\LK}[1]{{\color{red}{\bf LK: #1}}}

 \newcommand{\mk}[1]{{\color{magenta} #1}}
\else
 \newcommand{\MK}[1]{}
 \newcommand{\mk}[1]{#1}
\fi

\newcommand{\quotebox}[1]{\begin{center}\fcolorbox{white}{blue!15!gray!15}{\begin{minipage}{1\linewidth}\vspace{1pt}\center\begin{minipage}{1\linewidth}{\space\Huge``}{#1}{\hspace{1.5em}\break\null\Huge\hfill''}\end{minipage}\smallbreak\end{minipage}}\end{center}}


 \newcommand{\tick}{\ding{51}}
  \newcommand{\cross}{\ding{55}}

\newcommand{\gcol}[1]{{\bf \fontsize{3.5}{42}\selectfont \color{citecolor!80}(#1)}}
\newcommand{\rcol}[1]{{\bf \fontsize{3.5}{42}\selectfont \color{lightred!180}(#1)}}
\definecolor{citecolor}{RGB}{34,139,34}
\definecolor{lightred}{RGB}{241,140,142}

\newcommand\secvspace{\vspace{0cm}}
\newcommand\eqvspace{\vspace{0cm}}
\newcommand\figvspacetop{\vspace{0cm}}
\newcommand\figvspace{\vspace{0cm}}
\newcommand\tabvspace{\vspace{0cm}}
\newcommand\figcapvspace{\vspace{0cm}}

\begin{abstract}
    
State-of-the-art Vision-Language Models (VLMs) ground the vision and the language modality primarily via projecting the vision tokens from the encoder to language-like tokens, which are directly fed to the Large Language Model (LLM) decoder. 
While these models have shown unprecedented performance in many downstream zero-shot tasks (\eg image captioning, question answers, \etc), still little emphasis has been put on transferring one of the core LLM capability of In-Context Learning (ICL). 
ICL is the ability of a model to reason about a downstream task with a few examples demonstrations embedded in the prompt. 
In this work, 
through extensive evaluations, we find that the state-of-the-art VLMs somewhat lack the ability to follow ICL instructions. 
In particular, we discover that even models that underwent large-scale mixed modality pre-training and were implicitly guided to make use of interleaved image and text information (intended to consume helpful context from multiple images) under-perform when prompted with few-shot demonstrations (in an ICL way), likely due to their lack of \textit{direct} ICL instruction tuning. 
To enhance the ICL abilities of the present VLM, we propose a simple yet surprisingly effective multi-turn curriculum-based learning methodology
with effective data mixes, leading up to a significant $21.03\%$ (and $11.3\%$ on average) ICL performance boost over the strongest VLM baselines and a variety of ICL benchmarks. 
Furthermore, we also contribute new benchmarks for ICL evaluation in VLMs and discuss their advantages over the prior art.
\end{abstract}

\secvspace
\section{Introduction}\label{sec:intro}
\secvspace
A little more than a year ago, with the release of ChatGPT\footnote{\href{https://chat.openai.com/}{https://chat.openai.com/}} in late November 2022, Large Language Models (LLMs) made their historical debut showing, for the first time, that an artificial neural network can encompass in its parameters a potentially human-like understanding of language, essentially in all aspects of its distribution complexity including knowledge \cite{mmlu}, reasoning \cite{lewkowycz2022solving}, context understanding \cite{lampinen2022can} and other core capabilities \cite{seed,zhao2022vl}. 
\begin{figure}[t!]
    \centering
    \includegraphics[width=0.3\linewidth]{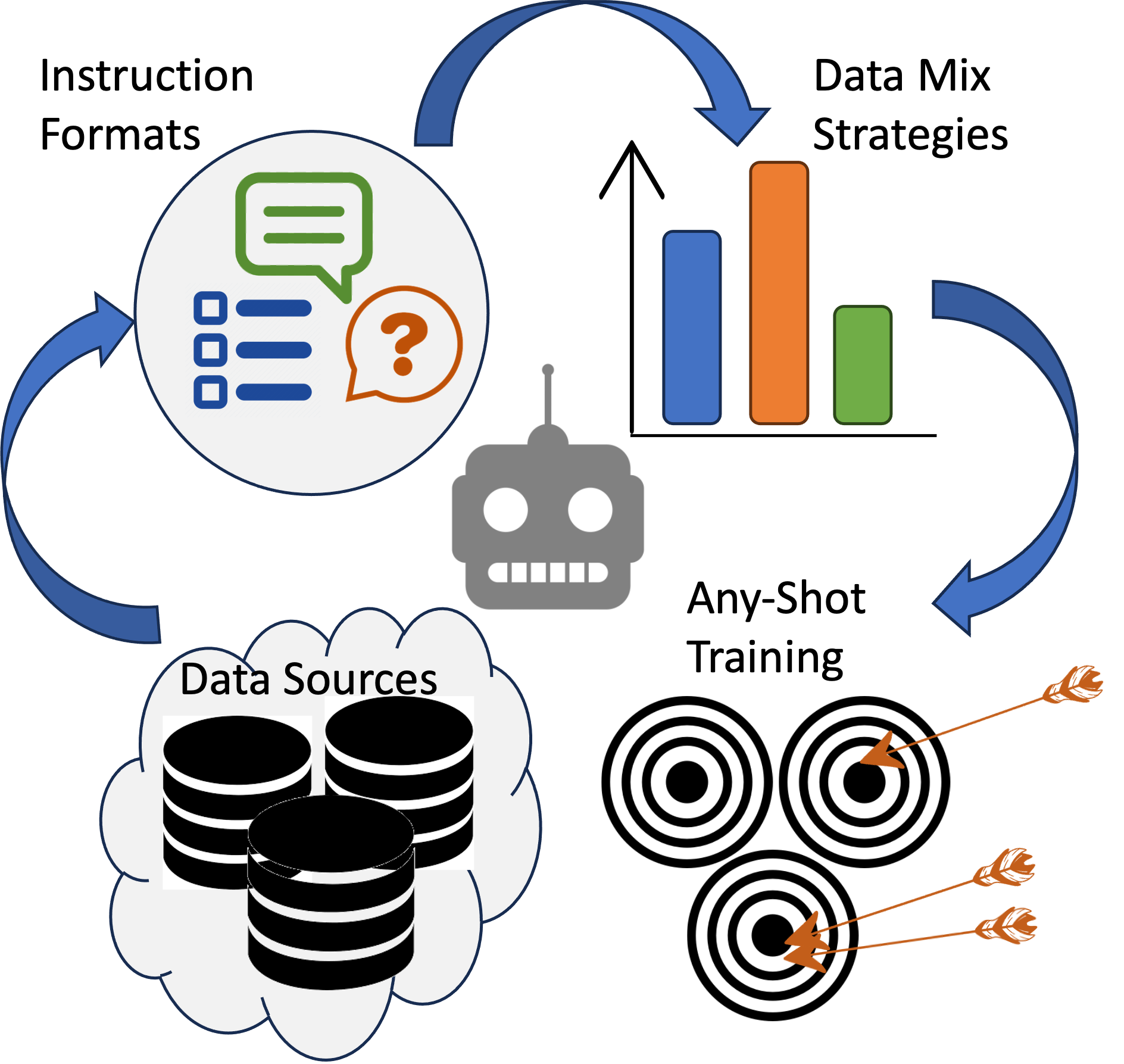} 
\caption{Multiple data sources are used to generate multi-modal ICL instructions varying the types of ICL tasks and type of semantic concepts shared within each instruction, teaching the VLM to properly correlate information between ICL in-context shots. Our insights on the best training data mix along with our proposed ``any-shot" training paradigm enhance the VLM's ICL abilities.}
\label{fig:high-level}
\figvspace
\end{figure}

Besides many other exciting LLM capabilities and discovered emerging properties, perhaps one of the more useful aspects of LLMs is their `foundation modeling' aspect - being able to understand and respond to language input, they are essentially `open' models and are not limited to any set of pre-defined tasks prescribed by the model training. 
In other words, as part of a typical use - one can explain a task to the model and have it comply (zero-shot inference), or if performance is not sufficiently satisfactory, one can provide a set of \emph{In-Context} demonstrations, illustrating to the model the desired task and input/output format (structure) via a few examples. The latter emerging property of LLMs is commonly referred to as In-Context Learning (ICL) or Few-Shot Learning (FSL) and is extremely useful in situations where more fine-grained control over the downstream task is needed. 

Motivated by the great advances of LLMs in language modeling~\cite{gpt3,vicuna2023,llama,touvron2023llama,mixtral}, there has been a huge interest in the community towards fusing these LLMs with other modalities.
Notably, these LLMs have recently been \textit{fused} with the vision modality~\cite{llava,idefics,minigpt,flamingo,minigptv2,llava-next}, Audio \cite{ltu,lyu2023macaw}, Speech \cite{whisper}, Documents \cite{wang2023docllm}~\etc.
The resulting models,~\eg~the recent Vision and Language Models (VLMs), have demonstrated parallel capabilities to ones demonstrated for language - namely an ability to handle arbitrary downstream tasks in a zero-shot manner, by only explaining the task to the model. 
While many Vision and Language (VL) benchmarks have been proposed \cite{mmlu,seed,zhao2022vl}, they mostly focus on zero-shot capabilities and tasks defined by common natural language terms, and with few exceptions~\eg~\cite{seed}, do not test the capability of VLMs to leverage paired, image+text, in-context demonstrations for the downstream task definitions. Moreover, as demonstrated by our findings (in \cref{sec:results}), on an exhaustive set of ICL evaluations, some contributed by our work, even the most advanced and recent VLMs \cite{llava,idefics,emu2,openflamingo,llava-next} still struggle with ICL. 
This trend seems to be consistent for both the \textit{leading visual encoder to LLM decoder} alignment methods \cite{llava,llava-next}, as well as for `encoder-free' techniques that directly processing a mixed stream of multi-modal tokens by a single transformer~\cite{openflamingo,emu2,idefics}.
Recent~VLMs~\cite{emu2,idefics,openflamingo} show some gains on downstream tasks, which can be solved with semantically unrelated in-context demonstrations such as answers to unrelated VQA questions on other images.
However, these SOTA VLMs commonly fail when explicitly challenged with tasks (\eg fine-grained few-shot visual recognition or instance recognition) completely defined by the ICL demonstrations (\eg an n-way episode of a few-shot visual recognition task). 
Arguably, 
ICL is one of the central capabilities of foundation models of all kinds (LLMs and VLMs alike) allowing for fine-grained control over downstream tasks. 
Hence, such shortcomings of leading VLMs need special attention.

In this work, we conjecture that the ICL performance of modern VLM approaches can be significantly improved by simple modifications to their training strategies, explicitly incorporating \emph{semantically-coherent} ICL tasks into their visual instruction tuning phase. 
We show that leveraging the~\textit{human-assistant} turn-based conversation structure common to visual instruction tuning \cite{llava,llava-next}, while extending it to a multi-image conversation, provides a simple and convenient vehicle for multi-shot explicit ICL training. In the multi-turn conversation format, the standard Causal Language Modeling (CLM) objective trains the model to operate in \textit{any-shot} scenario, that is later able to accept any number of in-context demonstration `shots' at inference time. The multi-turn conversation format also includes a `zero-shot' turn (the first turn of the conversation that under CLM object does not have any demonstrations in its processing context) thus providing replay support for zero-shot tasks and avoiding forgetting of the VLM's core capabilities. Equipped with this adaptation of the visual instruction tuning, we explore and provide insights on the most effective data mixing strategies via forming semantically coherent ICL tasks from multiple available data sources, guided by the requirement to have a common semantic aspect shared by all in-context demonstrations and currently trained query alike. Notably, even the approaches that are trained with mixed, multi-modal, tokenized stream \cite{openflamingo,emu2,idefics} containing multiple images in the same stream, under-perform (as shown in \cref{sec:results}) the above simple training technique proposed by us on top of the more light-weight visual instruction tuning alignment of \cite{llava}. As we conjecture, this is likely due to a lack of explicit ICL semantic coherence in their training data design.

To summarize our contributions are as follows: (i) We design a simple and yet surprisingly effective ICL visual instruction tuning strategy that can be easily added to standard visual instruction alignment tuning, significantly enhancing the explicit ICL capabilities of the VLM without forgetting its core zero-shot capabilities; (ii) We analyze and report insights on the most effective data mixes for our proposed ICL instruction tuning; (iii) We offer a set of ICL benchmarks that can assist in testing the ICL abilities of the present-day VLMs and can also act as a standard benchmark for the future.  
\secvspace
\section{Related Work}\label{sec:related}
\secvspace
We first provide an overview of zero-shot vision-language foundation models and then describe the literature closer to our line of work,~\ie~studying in-context learning in the domain of VLMs. 
\secvspace
\paragraph{Vision-Language Foundation Models:} 
Recently, VLMs have been adopted as the default choice for \emph{train once and use everywhere} paradigm, and have shown unprecedented performance for many vision-language understanding tasks,~\eg zero-shot classification, visual question-answering (VQA), image captioning, and many more. 
VLMs can be divided into two families of methods. 
One family of methods relies on dual-encoders (vision and text encoder) and usually trains the encoders with a contrastive objective by using a large corpus of paired image-text data scraped from the web.
Some representatives of this family of methods are CLIP~\cite{clip} (the first large-scale vision-language model), ALIGN~\cite{align}, OpenCLIP~\cite{openclip}, SigLip~\cite{sigclip} and MetaCLIP~\cite{metaclip}.
Furthermore, some methods have focused their attention on filtering noisy captions (\eg BLIP~\cite{blip}), employing textual nearest-neighbors~\cite{declip} or relying on using geometrically consistent representations~\cite{cyclip}, and caption augmentations~\cite{svlc,dac} for improving compositional reasoning aspects of these VLMs. 
In parallel, other methods have employed few-shot supervision~\cite{coop,cocoop,maple}, and also label-free finetuning~\cite{maxi,lafter,tap,alfassy2022feta}   .
The other group of methods aligns the visual modality with a frozen LLM. 
BLIP-2~\cite{blip2} bridges the modality gap between a pre-trained visual encoder and an LLM by using a Querying Transformer.
Instruct-BLIP~\cite{instructblip} proposes to improve~\cite{blip2} by employing instruction tuning. 
MiniGPT~\cite{minigpt} grounds a vision encoder with a frozen LLM (Vicuna~\cite{vicuna2023}) by only using a trainable linear projection layer between the two.
MiniGPT-V2~\cite{minigptv2} replaces the LLM with Llama-2~\cite{touvron2023llama} and enhances the performance by also
training and finetuning the LLM decoder. 
Llava~\cite{liu2023visual} also grounds an LLM with a pre-trained visual encoder and also proposes Visual Instruction Tuning, by carefully curating instruction-response pairs, to enhance the performance.  
The base Llava is further enhanced in Llava-1.5~\cite{liu2023improved} by careful curation of data and Llava-1.6~\cite{llava-next} also improves the previous version by incorporating some design changes and also modifying the instruction tuning data. 
Some other works~\cite{liu2023visual,wang2023visionllm,peng2023kosmos,chen2023shikra,bai2023qwen} also explore similar ideas.
Although these powerful and versatile encoder-decoder models can solve many tasks efficiently, their ability to learn or adapt to new tasks by only seeing a few contextual examples, instead of relying on a huge corpus of training data, is still under-explored. 
In our work, we take steps towards unlocking the ICL ability of these VLMs by a simple yet effective methodology of carefully curating ICL-specific data and altering the learning framework such that these models can become efficient in-context learners.  

\secvspace
\paragraph{In-context Learning for VLMs:}
In the natural language processing (NLP) literature, the ability of LLMs to solve novel tasks by only consuming a few demonstrations of the downstream task of interest has been formalized as in-context learning (ICL)~\cite{icl}.
For NLP, many different methods have been proposed to elicit the ICL ability in the powerful autoregressive models for many downstream tasks.
Notably~\cite{brown2020language} popularized few-shot learning for LLMs.
Similarly, other methods follow the basic idea of few-shot learning but achieve the goal in different ways,~\eg by breaking down a complex set of instructions in simpler steps~\cite{cot,treeofthought}.
Recently, the vision-language community has also shown considerable interest in ICL. 
Flamingo~\cite{flamingo} showed that ICL can scale up to large-scale vision language models. 
In particular, they improved downstream tasks like image captioning by only requiring a few examples. 
Flamingo's ability was unlocked by their novel fusion of visual information with the textual tokens. 
Recently, Emu2~\cite{emu2} showed that ICL for VLMs can be enhanced by scaling up the encoder-decoder models in modern VLMs with auto-regressive training. 
Similarly, Idefics~\cite{idefics} also scales up the vision (encoder) and language (decoder) to $80$ billion and shows effective ICL learning ability. 
Recent methods have made interesting progress toward obtaining better in-context learning by scaling the model size and specifically training for this purpose~\cite{emu2}. 
However, we ask the question: can we take an off-the-shelf VLM (like Llava~\cite{llava-next}) and convert it to an effective few-shot learner in an ad-hoc fashion? 
We answer this question in the affirmative and detail our approach in the following sections.

\secvspace
\section{Method}\label{sec:method}
\secvspace
In this section, we explain the proposed ICL-instruction-alignment approach. 
For ease of assimilation, we divide its description into 3 parts as illustrated in \cref{fig:high-level}.
\cref{sec:multiturn} discusses the details of our ICL alignment framework implemented as multi-turn ICL conversations inside the Llava~\cite{llava} visual-instruction alignment. 
\cref{sec:inst_format} explains the different ICL instruction task types used in our ICL alignment instruction sets and mixes. 
\cref{sec:data_sources} discusses the sources of data we used to construct semantically coherent ICL instructions that, as opposed to mixed multi-modal jointly tokenized internet data used in other works \cite{openflamingo,emu2,idefics}, is designed to guarantee the existence of a semantic concept shared between each set of ICL demonstrations (shots) and the ICL query. 
Finally, \cref{sec:icl_benchmarks} briefly discusses additional ICL evaluations contributed by us for evaluating `open vocabulary few-shot visual recognition' - one of the most common ICL tasks, somewhat neglected by the previous ICL benchmarks.
\begin{figure}[t]
    \centering
    \includegraphics[width=0.9\textwidth]{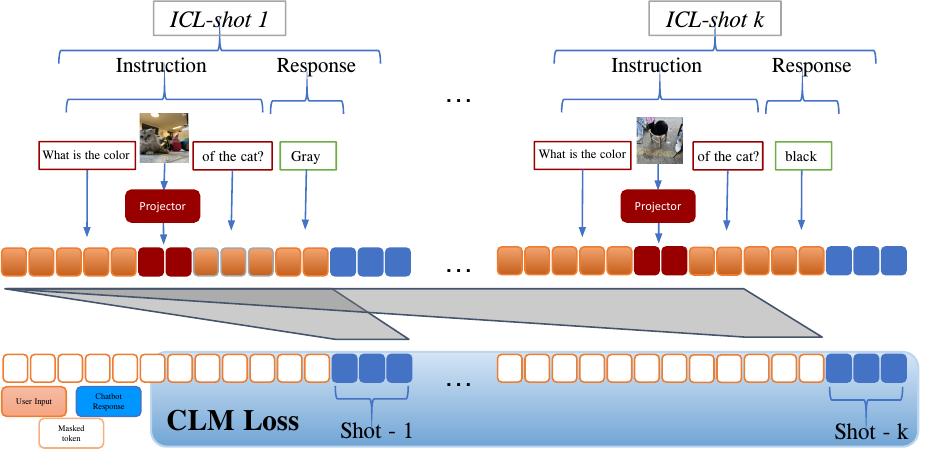}
    \caption{Causal (left only) attention and formatting the ICL examples as consecutive conversation turns, results in `any-shot' training where the first turn prediction is ``zero-shot", the next turn predicts the response given the context of the first, and so on, resulting in a dynamic ``any-shot" context. The grey shades illustrate the context that each turn's response attends to. As \cite{llava+} we do completion-only training, masking all but the desired responses (blue) in the target.}
    \label{fig:detailed}
    \figvspace
\end{figure}

\secvspace
\subsection{Multi-turn ICL conversations}\label{sec:multiturn}
\secvspace
A common strategy for aligning different pre-trained other-modality encoders (for visual, audio, speech, \etc) to an LLM is via multi-modal instruction tuning and associated training curriculum \cite{llava,llava+,ltu,llava-next,instructblip,minigpt,minigptv2}. 
Specifically, we build on the alignment model architecture of \cite{llava+} which consists of a large-scale pre-trained modality encoder $\mathcal{E}$, a modality projector $\mathcal{P}$, and an LLM decoder $\mathcal{D}$. 
The training data consists of `multi-modal' conversations between a `human' and a `gpt-assistant', the conversation is typically interleaving these two roles. The human role may contain modality tags (\eg \emph{<image>} in \cite{llava+}), yet in practice \cite{llava+} only implements this tag once (with a single image per conversation context) added only in the first (human) conversation role text. Therefore, for \cite{llava+}, the training samples are formed as:
\begin{equation}
\eqvspace
    \mathcal{D}(T_1 \oplus \mathcal{P}(\mathcal{E}(I)) \oplus T_2)
\eqvspace
\end{equation}
where $T_1$ and $T_2$ are tokenized parts of the conversation texts that come before and after the \emph{<image>} tag respectively, and $\oplus$ stands for concatenation. Typically, the training curriculum consists of two stages: (i) pre-training - freezing the modality encoder $\mathcal{E}$ and the LLM decoder $\mathcal{D}$ and training the projector $\mathcal{P}$ from scratch; and (ii) fine-tuning - both the projector $\mathcal{P}$ and the LLM decoder $\mathcal{D}$ jointly, keeping $\mathcal{E}$ frozen. In \cite{llava+} stage (i) training is comprised of short, single-turn (one user, one gpt) conversations, all para-phrasing a request by the user to describe a given image. At the same time, stage (ii) comprises more diverse multi-turn conversations combining different image task instructions and responses, albeit all relative to a single provided context image. 
In both phases of training, all the user input tokens, \ie~the instruction, and the aligned image tokens, $\mathcal{P}(\mathcal{E}(I))$, are masked 
and only the gpt-assistant's responses are used as target labels to train the alignment model with the standard Causal Language Modeling (CLM) objective. \cref{fig:high-level} visually describes the process.

We build our ICL visual instruction tuning as a simple and direct extension to \cite{llava+} alignment tuning. We keep the same architectural components, namely: $\mathcal{E}$, $\mathcal{P}$ and $\mathcal{D}$. We keep the pre-training stage (i) intact, yet extend (or replace) the fine-tuning stage (ii) train data with a mix of semantically-coherent ICL instructions. The discussion on the ICL task types and the sources of semantic coherence is covered in later \cref{sec:inst_format} and \cref{sec:data_sources}, while the format of the ICL instructions is as follows:
%
\begin{align*}
\eqvspace
&\text{Human: $S_1^1$<image>$S_1^2$}~~\text{GPT: $R_1$}\\
&\text{Human: $S_2^1$<image>$S_2^2$}~~\text{GPT: $R_2$}\\
&\text{Human: $S_3^1$<image>$S_3^2$}~~\text{GPT: $R_3$}~~\cdots
\eqvspace
\end{align*}
accompanied with a corresponding ordered list of images $[I_1, I_2, I_3, \cdots]$. Here each $\{S_j^1, S_j^2, I_j, R_j\}$ comprises an ICL `shot', that is a single in-context demonstration example. Such multi-turn ICL instruction format required only minor modifications to the official \cite{llava+}~code\footnote{\href{https://github.com/haotian-liu/LLaVA}{https://github.com/haotian-liu/LLaVA}}. 
The only modification was to add multi-image input support in training as opposed to the original \cite{llava+} training phases, which only used single-image conversations. 
Interestingly, in combination with input masking of human turns,
the CLM objective, and the CLM attention - allowing tokens to attend only to their left, our simple ICL conversation format becomes an `any-shot trainer'. 
Indeed, the first shot has no other shots in its attended context and therefore serves as a `zero-shot' instruction replay - effectively reminding the aligned model that it needs to continue supporting this mode of operation, as aligned with~\cite{llava+}. 
Then, each shot number $i$, observes the shots $1,\cdots,(i-1)$ in its attended context and hence trains the model to support $(i-1)$-shot ICL instructions.

\secvspace
\subsection{ICL instruction task types}\label{sec:inst_format}
\secvspace
We set each ICL instruction shot  $\{S_j^1, S_j^2, I_j, R_j\}$ to one of the following instruction-response formats: (a) Open QA - the texts $S_j^1$ and $S_j^2$ form an open question, where image $I_j$ tokens are embedded in the semantically valid position prescribed by the question language context. The desired response $R_j$ is formulated as natural language text (commonly a single sentence); (b) Multiple-choice QA - $S_j^1$ is empty and $S_j^2$ contains a question followed by several answer options marked by A, B, C, ..., $R_j$ contains a single letter correct answer choice; (c) Captioning - $S_j^1$ contains a para-phrased in different ways `describe the image' request, $S_j^2$ is empty, and $R_j$ contains the image description. For all ICL instruction task types, the shots in the same instruction have a semantic coherence in the form of a semantic concept shared across all the shots in the ICL instruction (e.g., all questions ask about a certain type of object attributes such as color, or all captions share a common style and/or intent - e.g. describe locations of objects on the image). The coherence is achieved via careful data curation explained in \cref{sec:data_sources}. Intuitively, in a single ICL instruction, all the shots are of the \textit{same} instruction task type (a) Open question answering (QA), (b) multiple choice (MC), or (c) Captioning (Cap), respectively. More details and examples of each ICL instruction type are provided in Supplementary.

\secvspace
\subsection{Data sources for ICL instruction mixes}\label{sec:data_sources}
\secvspace
As we show in our ablation studies in \cref{sec:ablations}, correct mixing of ICL instruction types, as well as sources of semantic coherence, is crucial for maximizing ICL instruction alignment performance. In other words, the ICL curriculum is an important component of our approach. To induce semantic coherence between shots of each ICL instruction, that is, the presence of some semantic concept shared between all shots in each any-shot ICL conversation explained in \cref{sec:multiturn}, we collect our ICL instructions from the following data sources: SEED benchmark \cite{seed} partitions 1-5 (Scene Understanding, Instance Identity, Instance Attributes, Instance Location, Instances Counting) and VL-Checklist \cite{zhao2022vl} 13 partitions (\eg~Color, Material, Action, Object, Positional Relation, Action Relation, Action Attribute). 
In all cases, we use the source dataset partition to sample the \textit{k}-shots of each ICL instruction from the same partition. As described in \cref{sec:inst_format} all shots of the ICL instruction are then structured in the same format according to one of the 3 formats described. As extensively analyzed in \cref{sec:ablations}, the data mixes of instruction formats and sources of semantic coherence (the aforementioned dataset partitions) have a crucial importance on downstream ICL performance and generalization. We contribute insights into these empirical observations on the optimal data curriculums (resulting from significant investments in computing) with hopes of inspiring exciting future research direction of enhancing ICL in VLMs. 
Additionally, we found that aside from the first (zero-shot) turn replay, adding the original \cite{llava+} fine-tune data portion to the training curriculum for providing additional replay support is useful.

\secvspace
\subsection{ICL benchmarks}\label{sec:icl_benchmarks}
\secvspace
With few exceptions (e.g., task 23 in Seed-2 \cite{seed2}), most of ICL evaluation benchmarks so far were constructed ad-hoc from largely unrelated $k$-shot episodes randomly sampled from either a VQA dataset (VQAv2, VQA, OKVQA, TextVQA, VizWiz, \etc) or a visual dialog dataset \cite{visdial}, as evaluated by \cite{idefics,emu2}. While such evaluations demonstrate improvements with adding more shots, the lack of semantic relations between the in-context demonstration shots and the query makes it unclear whether most of the improvement does not come from, e.g., matching the desired output format. Additionally, practical ICL use cases are often parallel to the now classical problem of few-shot learning, particularly few-shot visual recognition. Equipped with these insights, we formulate these additional ICL benchmarks for VLMs: (i) Fine-grained few-shot visual recognition ICL benchmarks derived from popular few-shot tasks: Stanford Dogs~\cite{dogs}, CUBS~\cite{cubs200}, FOOD-101~\cite{food}, Stanford Cars~\cite{cars}, Flowers~\cite{flowers}; (ii) `unseen' Seed tasks 6-8 reformulated into ICL episodes as described above (but not used for training); (iii) put aside validation portion of `seen' Seed task 5 (instance counting) as the hardest of the training Seed tasks requiring all the other capabilities of the training Seed tasks 1-5 to execute; (iv) put aside validation portion of `seen' VL-checklist partitions. We detail the exact statistics of all these datasets and splits in the Supplementary.
 \secvspace
\section{Results}\label{sec:results}
\secvspace
 In \cref{sec:eval_settings}, we first provide an overview of the different datasets used in our work, followed by a brief description of the baselines and an explanation of the implementation details. We discuss our main findings in \cref{sec:main_findings}, uncovering the strong potential for our proposed ICL instruction tuning for decoder VLMs, leading to over $11\%$ average absolute improvement over the strongest baseline. We provide an ablation study in \cref{sec:ablations} digging deeper into the different aspects of our method, primarily focusing on the properties of the ICL instruction tuning data mixes that lead to the aforementioned improvements. Additionally, we highlight the scaling potential of our approach by increasing the ICL instruction tuning data size. Finally, we ablate the importance of replaying non-ICL data instruction tuning data, concluding that (a) due to non-ICL instructions replay, our model avoids forgetting the core capabilities of the base model \cite{llava+} as measured by the standard and extensive MME \cite{mme} benchmark; (b) the replay visual instruction tuning data generally improves our model's ICL performance.

\secvspace
\subsection{Evaluation Settings}\label{sec:eval_settings}
\secvspace
\paragraph{Datasets:}
We briefly list the datasets used to form our ICL-instruction tuning mixes and/or for evaluation of ICL or other capabilities of the resulting model and baselines.

\begin{enumerate}
    \item \textbf{SEED-Bench-2} \cite{seed2} - 
    SEED-Bench-2 \cite{seed2} is an extended version of SEED-Bench \cite{seed} that features 
    a total of 27 evaluation dimensions. Our use of SEED-Bench-2 is two-fold. We use $100\%$ of data of tasks $1$-$4$ 
    (scene and `Instance' tasks)
    to form ICL instructions of multiple-choice type (SEED-Bench format). We use $90\%$ of data of tasks $5$ (`Instances Counting', arguably hardest of $1$-$5$ tasks) to form multiple-choice ICL instructions and its $10\%$ to form $2$-shot multiple-choice ICL `Instances Counting' evaluation. Additionally, we use $100\%$ of tasks $6$-$8$ (relation, interaction, reasoning) for $2$-shot multiple-choice ICL evaluation of `Unseen SEED-tasks'. Additionally, we evaluate directly on SEED-Bench-2 task 23 - `In-context captioning', the only SEED ICL task. We provide more details on ICL with SEED-Bench-2 in the Supplementary. 

    \item \textbf{VL-checklist} \cite{zhao2022vl} - is a benchmark constructed from Visual Genome~\cite{visualgenome}, SWiG~\cite{swig}, VAW~\cite{vaw}, and HAKE~\cite{hake}. 
    Overall, VL-checklist has 13 such partitions according to a shared semantic aspect of MM evaluation: attribute, relation, or object, each further subdivided. We split VL-checklist into two non-overlapping subsets, using $70\%$ for ICL instruction tuning and $30\%$ for ICL $2$-shot testing. For VL-checklist we form both multiple-choice and open QA ICL task types for all partitions and, additionally, ICL captioning task types for 6 partitions: Color, State, Material, Size, and Action attribute/relation. More details and examples are provided in the Supplementary.
    
    \item \textbf{LLaVA visual instruction tuning dataset} \cite{llava+} - dataset of 655K visual instructions constructed by the LLaVA team \cite{llava} as part of their visual instruction tuning works \cite{llava,llava+,llava-next}. Built from a combination of COCO, GQA, OCR-VQA, TextVQA, and VisualGenome data.
    We do not use this data for ICL training, only for general visual instructions replay intended for preserving the general capabilities of the base \cite{llava+} VLM which we build upon.
    \item \textbf{Stanford Dogs} \cite{dogs} - is a dataset of 120 dog breeds with  ~150 images per class, 12,000 images for training and 8,580 for testing.
    \item \textbf{CUB} \cite{cub} - contains 200 images of different types of bird species with 5,994 samples for training and 5,794 for testing. The annotations include several attributes and localization. 
    
    \item \textbf{Flowers} \cite{flowers} - consists of 102 images of flower categories common in the UK. Each class contains between 40 and 258 images. Its test set has 6149 images.
    
    \item \textbf{Food-101} \cite{food101} - is a data set of 101 food categories, with  ~100K images. Each class contains 250 manually reviewed test images and 750 training images. The full test set includes 25250 images.
    
    \item \textbf{Stanford Cars} \cite{cars} - consists of 196 classes of cars with a 50/50 division to train and test. Categories are typically at the level of Make, Model, and Year. Test set contains 8041 samples.

\end{enumerate}
Datasets 4-8 are fine-grained few-shot datasets that are completely unseen during the training of our model and are used for evaluating our ICL instruction tuning generalization testing only. We test them in a 2-way / 1-shot mode, meaning that for each testing episode (ICL task instance), in addition to the test image, we sample a random image from the same class as the test image and a random image from a different class than the test image. We then provide them to the model with their respective labels, and finally, we provide the model with the test image as a query asking the model to choose among the two class possibilities. We use only the test partitions of these datasets and generate an episode for each test sample from the test sets. We generate an average of 10763 episodes from each dataset.

\secvspace
\paragraph{Baselines:} We compare with the following state-of-the-art baselines. 
\begin{enumerate}
    \item IDEFICS 9B~\cite{idefics} - is an open-source reproduction of Flamingo~\cite{flamingo},  trained on a large corpus of publicly available datasets, and shows strong in-context learning abilities.
    The architectural details are kept similar to~\cite{flamingo}; however, the total number of model parameters differ. 
    \item OpenFlamingo \cite{openflamingo} - also proposes to provide an open-source alternative to the original Flamingo~\cite{flamingo}, and sticks to the design proposed by~\cite{flamingo}, while training on different data than the original Flamingo. 
    \item EMU2 \cite{emu2} - is a generative multimodal model capable of generating both images and texts. It has good ICL abilities and has been shown to achieve better performance than IDEFICS-80B and Flamingo-9B on multiple tasks.
    \item LLaVA-1.5 13B \cite{llava+} - builds upon the framework proposed by the original LLaVA~\cite{llava} but makes some design changes, like using a more expressive projection between the vision and the language tokens and also trains on different instruction-tuning data. 
    \item LLaVA-next (1.6) 13B \cite{llava-next} - improves LLaVA-1.5 by altering the instruction tuning data and the visual processing pipeline. 
\end{enumerate}

\secvspace
\paragraph{Metrics:} All experiments, unless stated otherwise, were measured using the accuracy of the models' generation abilities, requiring an exact match of the GT and the predicted model. The accuracy is calculated as the percentage of exact-match responses. The response may be a single character, as in the multi-choice or few-shot scenarios, or a full string in the captioning and open-question tests. An exception to the exact-match criteria is the SEED-23 task, where the output is measured using the perplexity value, selecting the most likely character out of the four choices (the choices are provided as part of task 23; one of the choices gives the desired caption that should be selected given the context of 2 image examples and their respective semantically related captions). 

\secvspace
\paragraph{Implementation Details:}
We build upon the \cite{llava+} codebase and use its default training parameters when tuning our model. We extended the codebase to allow accepting a list of encoded images with each visual instruction during training (the original code supported only a single image tag and image input per visual instruction conversation). Following this extension, a training conversation is allowed to contain a number of \texttt{<image>} tags matching the length of the associated images list. These \texttt{<image>} tags are replaced according to the order of the images list. We use Vicuna-1.5-13B \cite{vicuna2023} LLM backbone and use the recommended by \cite{llava+} 2K context length, which supports up to 3-images conversations for our any-shot ICL instruction tuning (each image takes about 500 tokens of the context). Larger models and more extensive investment in computing would allow much larger context length and support for much longer any-shot sequences, which we believe would improve the performance further. We used a single 8x A100 80GB Nvidia GPU node for all of our training runs.

\secvspace
\subsection{Main findings}\label{sec:main_findings}
\secvspace
In this section, we discuss our findings from a comprehensive set of multi-modal ICL evaluation experiments carried out on a collection of held-out validation splits of our ICL instruction tuning mix datasets (SEED-Bench-2 task $5$ (Instance counting), VL-checklist partitions), completely unseen ICL tasks evaluating our method generalization (SEED-Bench-2 tasks $6$-$8$, 5 few-shot fine-grained visual recognition datasets), as well as native ICL captioning evaluation of SEED-Bench-2: task 23 (currently the only semantically coherent multi-modal ICL evaluation task).

\secvspace
\paragraph{Few-shot visual recognition evaluations:}
We evaluate our model and the baselines (discussed in \cref{sec:eval_settings}) on the collection of $5$ standard (fine-grained) few-shot datasets (listed in \cref{sec:eval_settings}) using $2$-way / $1$-shot episodes of multiple-choice ICL task type. All these datasets are unseen during training, and test our method's multi-modal ICL generalization ability and the baselines. The results of this evaluation are presented in \cref{tab:model_comparison_fs}. Clearly, our method was able to leverage the proposed multi-turn successfully conversation-based any-shot ICL tuning and the ICL data mix to generalize well to these unseen fine-grained few-shot visual recognition tasks over a diverse set of visual categories from the 5 datasets, including food, vehicles, animals and plants. On average, our method successfully improves by over $12\%$ over the top-performing baseline (\cite{llava-next}), notably improving over IDEFICS 9B \cite{idefics}, Otter\cite{otter} and EMU2 \cite{emu2} by over $25\%$ on average despite of their strong, mixed-modality, pre-training which includes multiple images in their respective training contexts. Intuitively, we attribute these large gains to a likely lack of explicit semantic coherence in the baselines' training data. Interestingly and expectedly, our semantically coherent ICL pertaining generalizes well to these tasks, showing good potential for further improvement by introducing our proposed ICL instruction tuning to all LLM-decoder-based VLMs and highlighting the importance of fixing the ICL performance in those otherwise very strong models.

\begin{table}[t]
\centering
\small 
\resizebox{0.65\textwidth}{!}{
\begin{tabular}{l|ccccc|c}
\toprule
Model & Food  & Cars & Dogs & CUB & Flowers & AVG \\
\midrule
IDEFICS 9B & 65.94  &77.30 & 50.93 &  62.00  & 55.29 & 59.30 \\
OpenFlamingo 9B &52.3&	57.43 & 50.47 & 51.2 & 48.78 & 52.04 \\
EMU2 37B & 59.92 & 55.42 & 50.27 & 53.56 & 52.76 & 52.47 \\
Llava-1.5 13B & 87.19 & 57.30 & 33.00 & 58.24 & 58.60 & 60.77 \\
Otter & 33.15 & 22.19 & 33.8 & 24.62 & 60.01 & 34.75 \\
Llava-1.6 13B & 89.15 & 84.70 & 72.39 & 67.90 & 65.58 & 72.96 \\
\midrule
Ours 13B & 97.44  & 96.51 & 79.85 & 78.67 & 76.51 & 85.79\\

\gcol{gain} &\gcol{+8.29}&\gcol{+11.8}&\gcol{+7.46}&\gcol{+10.77}&\gcol{+10.93}&\gcol{+12.83}\\ 
\bottomrule
\end{tabular}
}
\caption{Comparison across our proposed fine-grained few-shot ICL tasks. 
The last row highlights the gains compared to Llava-1.6 13B, which is the leading baseline.}
\label{tab:model_comparison_fs}
\tabvspace
\end{table}

\secvspace
\paragraph{Additional ICL evaluations:}
We present additional ICL evaluations in \cref{tab:model_comparison}. These are comprised of a mix of ICL validation set tasks (SEED-Bench-2 Instance Counting, multiple-choice / QA / captioning ICL on VL-checklist) and unseen (during training) ICL generalization tasks (SEED-Bench-2 tasks $6$-$8$ multiple-choice ICL and in-context captioning task 23). Similar findings to those observed for the fine-grained few-shot ICL experiments in \cref{tab:model_comparison_fs} are seen. Again, our method improves by over $10$ points beyond the top-performing baseline, further highlighting the importance of our contributions and the need to incorporate the proposed semantically coherent ICL training into LLM-decoder-based VLMs. Notably, our method improves over $4$ points in SEED-Bench-2 task 23 (in-context captioning) - the only semantically coherent ICL benchmark task to the best of our knowledge that was not contributed by us, further confirming our intuition above. We note, on the low results of the OpenFlamingo model, that in many cases, the model returns empty values, which adds up as mistakes.
On average, across all datasets, we see an average boost of $11\%$.

\begin{table}[t]
\centering
\small 
\resizebox{0.7\textwidth}{!}{\begin{tabular}{l|cccccc|c}
\toprule
Model & SEED 23 & Unseen & Ins Count & MC VL & QA VL & Cap VL & AVG\\
\midrule
IDEFICS 9B & 45.00 & 27.00 & 28.00 & 63.77 & 10.00 &  2.37 & 29.36 \\
OpenFlamingo 9B & 25.00 & 24.80 & 18.25 & 45.00 &0.77 &0.41 & 19.04\\
EMU2 37B & - & 44.15 & 30.9 & 67.60 & 12.00 & 2.25 & 31.38 \\    
Llava-1.5 13B & 44.17 & 61.00 & 60.00 & 33.00 & 1.56 & 5.62 &34.26 \\
Otter& 34.7 & 27.09 & 34.27 & 65.79 & 12.75 & 2.27 & 29.47\\ 
Llava-1.6 13B & 40.8 & 59.00 & 49.00 & 88.00 & 19.48 & 2.27& 43.09 \\

\hline
Ours 13B & 49.16 & 62.20 & 65.30 & 95.37 & 24.50 & 23.40 & 53.32 \\

\gcol{gain} & \gcol{+4.16} & \gcol{+3.20} & \gcol{+16.30} &  \gcol{+7.37} & \gcol{+5.02} & \gcol{+21.03} & \gcol{+10.23} \\
\bottomrule
\end{tabular}}
\caption{Evaluating ICL tasks built from SEED and VL-Checklist. MC=multiple choice, QA=open QA, Cap=ICL captioning. Ins Count=our SEED-2 task 5 val split. Unseen=our ICL test based on SEED-2 tasks 6-8. SEED 23=SEED-2 task 23. EMU2 did not report on SEED and their public code does not support perplexity inference.}
\label{tab:model_comparison}
\tabvspace
\end{table}
\secvspace
\section{Ablations}\label{sec:ablations}
\secvspace
In this section, we explore the different aspects and design choices that contribute to the success of our approach of ICL instruction tuning. In \cref{sec:data_mix} we explore the effects of different choices and mixes of \textit{task types} and \textit{shared semantic concepts} of the ICL instructions on the performance of the model. In \cref{sec:data_scaling} we discuss the scaling properties of our model with respect to adding more ICL data. In \cref{sec:replay} we explore how well our additional ICL instruction tuning preserves the base capabilities of the \cite{llava+} model we start from. In particular, we show that replaying non-ICL instruction data (from \cite{llava+}) helps preserve the model capabilities (as measured through the MME \cite{mme} metrics), and also, surprisingly, generally benefits performance. Finally, in \cref{sec:shots} we analyze the model's ability to effectively leverage in-context (visual + text) information during inference. We measure the model's performance by ranging the number of shots between $2$ and $0$, showing that as desired, our ICL-instructions-tuned model strongly benefits from the addition of more shots (in-context visual and text information).

\secvspace
\subsection{ICL Data mixes ablation}\label{sec:data_mix}
\secvspace
We evaluate multiple different ICL instruction data mixing strategies and the effect of including the base \cite{llava+} fine-tuning data into our mix. These experiments are summarized in \cref{tab:mixes}. We will refer to the different mixes via their mix ID. As we can seen from comparing mixes $1$ and $2$, and our best mix $5$ with mix $6$, including \cite{llava+} data has a positive effect on the performance of the ICL tasks (and also preserves the aligned model capabilities). We, therefore, include it in all the rest of the data mixing recipes. Next, we discuss the aspects of ICL instruction type and shared semantic concepts in the ICL instructions. We use the average across all evaluations from \cref{tab:model_comparison_fs} and \cref{tab:model_comparison} as our average accuracy measure. Here, we derive some important insights that significantly boost our ICL instruction tuning performance (as described below), yet of course, with more investment in compute performance could be significantly improved further. As our experiments already show great promise for our proposed ICL instruction tuning, we leave the investigation to future work.

\def\checkmark{\tikz\fill[scale=0.4](0,.35) -- (.25,0) -- (1,.7) -- (.25,.15) -- cycle;}

\begin{table}[t]
\centering
\resizebox{0.8\textwidth}{!}{\begin{tabular}{l|c|cccc|ccc|c}
\toprule
\shortstack{Mix\\ID}& \shortstack{LLaVA \\data} &\shortstack{Attributes} & \shortstack{Relations} & \shortstack{Categories} & {Instances} & \shortstack{Open\\Questions} & \shortstack{Multiple\\Choice} & {Captioning} & \shortstack{AVG} \\\midrule
1  & & 0.00\% & 0.00\% & 0.00\% & 100.00\% & 0.00\% & 100.00\% & 0.00\% &  56.56 \\ 
2  & \checkmark{} &0.00\% & 0.00\% & 0.00\% & 100.00\% & 0.00\% & 100.00\% & 0.00\% & 65.41 \\ 
3  & \checkmark{}& 66.67\% & 	0.00\% & 	0.00\% & 	33.33\% & 50.00\% & 50.00\% & 0.00\%  & 66.35 \\ 
4 &\checkmark{}& 75.00\% &	12.50\% &	0.00\% &	12.50\%  &  65.00\% & 5.00\% & 30.00\% & 68.44 \\ 
5 &\checkmark{}& 45.45\% &	15.15\% &	36.36\% &	3.04\%  &39.40\% & 42.42\% & 18.18\% & 69.33 \\ \bottomrule

6 &\checkmark{}&46.87\%&	15.62\%	&37.05\%	&0.00\%&40.625\%	&40.625\%	&18.75\%&68.10\\
7 & & 45.45\% &	15.15\% &	36.36\% &	3.04\%  &39.40\% & 42.42\% & 18.18\% & 67.84\\
\bottomrule
\end{tabular}}
\caption{Shared semantic concept (between ICL shots in the same any-shot instruction) mixing ablation (left). Instruction format mixes effect on model performance ablation (right). `LLaVA data' indicates if tuning data from \cite{llava+} was used in the mix.}
\label{tab:mixes}
\tabvspace
\end{table}

\begin{figure*}[t!]
\vspace{2em}
\begin{minipage}{0.32\textwidth}
\centering
\resizebox{0.90\textwidth}{!}{\begin{tabular}{ccc}
    \toprule
& Cognition Total & Perception Total\\
\midrule
Llava1.5 & 295.36 & 1531 \\
Ours & 301.43 & 1520\\
         \bottomrule
    \end{tabular}}
    \captionsetup{type=table}
    \caption{MME scores of baseline and our model.}\label{tab:mme}
\end{minipage}
\hfill
\begin{minipage}{0.25\textwidth}
\centering
\resizebox{0.8\textwidth}{!}
{\begin{tabular}{cccc}
    \toprule
& MC& QA & Cap\\
\midrule
Two-shot & 95.37 & 24.5 & 23.4 \\
One-shot & 87.9 & 12.9 & 0 \\
Zero-shot & 87.4 & 12. & 0 \\
         \bottomrule
    \end{tabular}}
    \captionsetup{type=table}
    \caption{Varying number of Shots on VL tasks.}\label{tab:shots}

\end{minipage}
\hfill
\begin{minipage}{0.32\textwidth}
\includegraphics[width=0.8\textwidth]{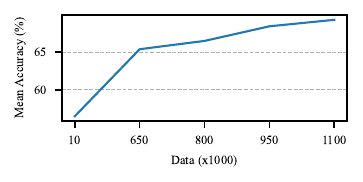}
\centering
    \vspace{-1.1em}
    \caption{Mean Accuracy (\%) while scaling ICL instruction data.}\label{fig:scaling}
\end{minipage}

\figvspace
\end{figure*}

\secvspace
\paragraph{Mixing ICL instruction formats:}
Here we examine the importance of the ICL instruction format in ICL tuning examples formation (\cref{tab:mixes} right side). Namely, we measure the relative importance of open questions vs multiple-choice vs captioning ICL instructions. Comparing mixes $2$ - $5$, starting from multiple-choice only ICL instructions, adding open questions ICL poses a significant benefit to performance. Additionally, we find that including ICL captioning task helps performance, and the best mix that is behind our best-performing model reported in \cref{tab:model_comparison_fs} and \cref{tab:model_comparison}, is reported under mix ID $5$.

\secvspace
\paragraph{Shared semantic concepts within ICL instructions:}
Part of our ICL instruction design is having a shared semantic concept (category, instance, relation, attribute, etc.) in all the shots of a single ICL instruction. This design teaches the model to be sensitive and be able to leverage the shared semantic information between shots. Here we analyze which types of semantic concepts composition in the mix is most beneficial. According to our findings (\cref{tab:mixes} left side), shared attribute concepts seem to be most beneficial, as long as it doesn't completely dominate the data. The second most important concept type is `object categories' (e.g., having cat notion in all shots), while interestingly `object instances' (e.g., asking about particular cat instances in all shots) seems less helpful to the best mix. Yet removing instances data completely hurts performance (mix 6). Relations shared concepts are represented in the third place, yet comparing mixes with and without them, we see they provide a significant boost.

\secvspace
\subsection{Data scaling}\label{sec:data_scaling}
\secvspace
Here we evaluate the scaling potential of our ICL instruction tuning approach. \cref{fig:scaling} presents our model average performance with scaling the ICL instruction tuning data. As can be seen, more ICL data consistently improves average performance over ICL tasks, ending with a positive gradient indicating the potential for additional improvement with further scaling.


\secvspace
\subsection{Preserving the base model abilities}\label{sec:replay}
\secvspace
One of the important questions is, with the addition of the ICL instructions - are we able to preserve the base capabilities of the aligned VLM? In particular, the \cite{llava+} we are starting from. We answer this question in the affirmative in \cref{tab:mme}, comparing our ICL instruction tuned model MME \cite{mme} scores with the base \cite{llava+} model. We attribute the preservation of the model's base capabilities to the importance of replaying the \cite{llava+} data, which was also found effective for boosting the ICL tasks performance in \cref{sec:data_mix}.




\subsection{Leveraging the shots information}\label{sec:shots}
Finally, in \cref{tab:shots} we test our model performance on a subset of our ICL evaluation tasks changing the number of in-context shots between $0$ and $2$ (the maximum allowed by the set 2K context length of \cite{llava+}). As we see, our model positively improves with the addition of more shots, as expected.
\secvspace
\section{Summary \& Conclusions}\label{sec:summary}
\secvspace
In this work, we analyzed and proposed some ways to improve the ICL tasks performance of Vision and Language Models. Our proposed approach leverages carefully designed ICL instructions and their respective data mixes, as well as the proposed any-shot training paradigm resulting in a model able to take advantage of in-context examples to better perform on a variety of tasks, such as multiple choice Q\&A, instance counting, captioning etc. Our work includes extensive comparisons to strong baselines. We also propose and evaluate few-shot visual recognition posed as ICL on multiple fine-grained datasets. 
Simple as it is, our approach shows significant and consistent gains across all evaluations, suggesting that future VLMs can significantly benefit from the proposed ICL instruction tuning (with shared semantic concepts) as well as from the any-shot training paradigm. We believe our ideas are orthogonal to our current implementation and can be easily re-used for many other models.
Our work suggests exciting future work directions including exploring ICL instruction tuning with longer LLM context enabling longer any-shot sequences, additional exploration of ICL instruction data mixes, additional ICL task types, and more ideas on possible shared ICL semantics.

\bibliographystyle{splncs04}
\bibliography{egbib}
\clearpage
\appendix

\section*{Appendix}
\setcounter{section}{0}

We first provide details about the training data and the ICL instruction tasks (Section~\ref{sec:ICL}), then list the detailed ICL results per task (Section~\ref{sec:icl_detailed_results}).
Later, provide details about the ICL instruction task types during the test phase (Section~\ref{ICL_test}) and finally conclude with qualitative visualizations.

\section{ICL instruction task types in training} 
\label{sec:ICL}

We construct our training data from two public datasets, namely SEED benchmark~\cite{seed} and VL Checklist~\cite{zhao2022vl}.
Next, we provide details regarding how the training data is formalized. 
\subsection{SEED benchmark}
\label{seed}
\begin{figure}[h]
    \begin{subfigure}{\textwidth}
        \centering
        \includegraphics[width=0.8\linewidth]{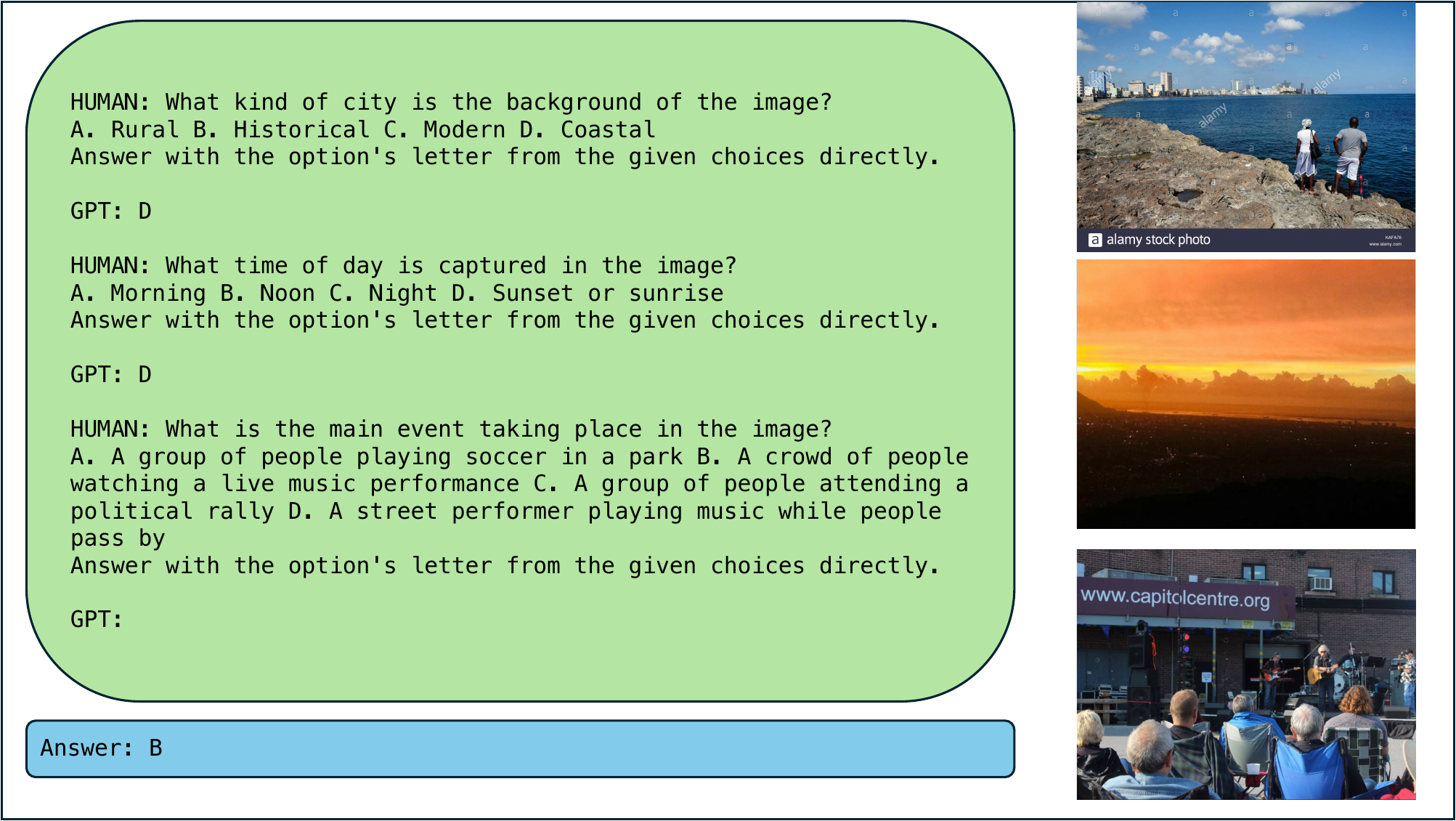}
        \caption{SEED benchmark task 1: Scene Understanding}
    \end{subfigure}
    
    \begin{subfigure}{\textwidth}
        \centering
        \includegraphics[width=0.8\linewidth]{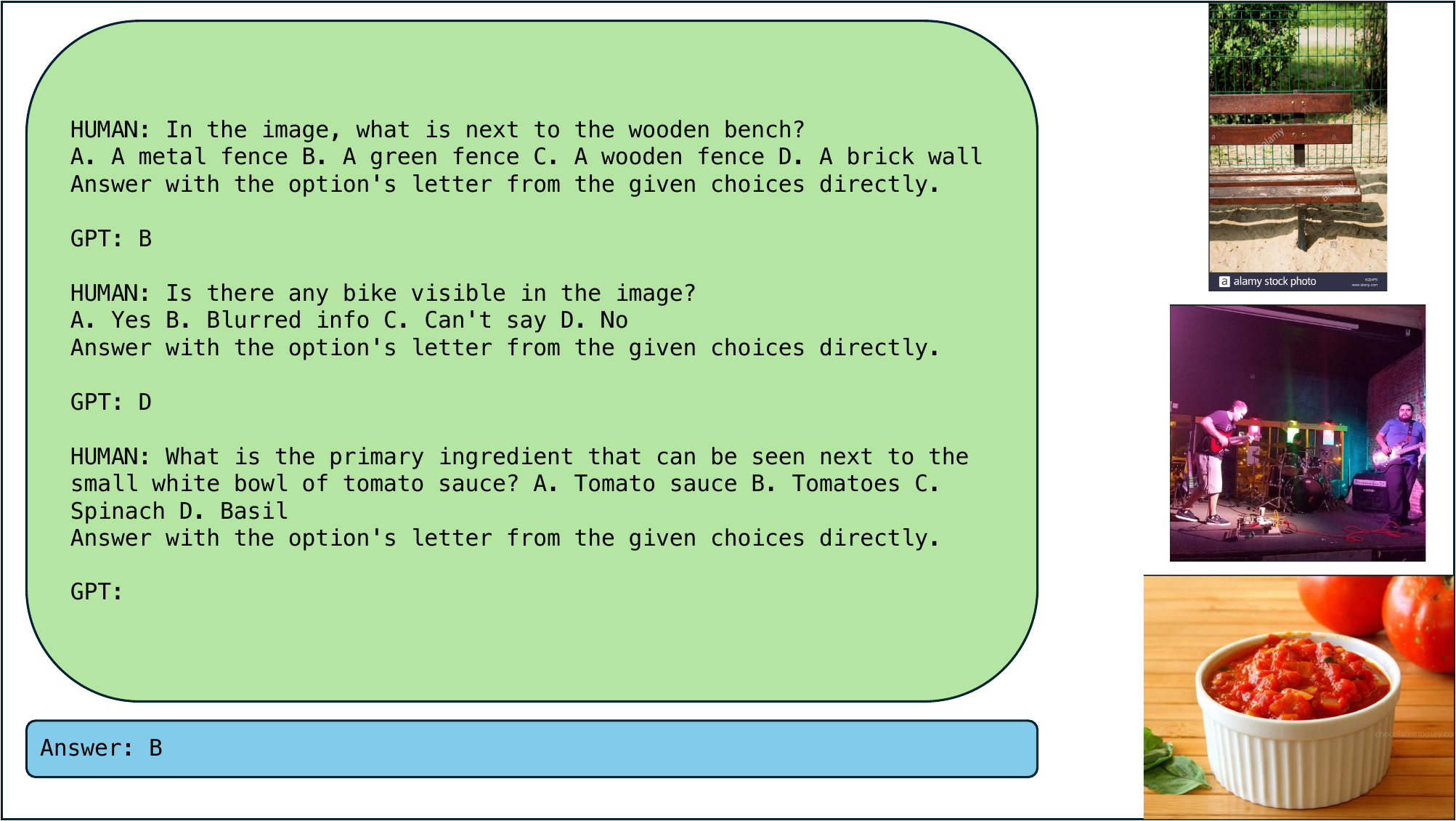}
        \caption{SEED benchmark task 2: Instance Identity}
    \end{subfigure}
    
    \begin{subfigure}{\textwidth}
        \centering
        \includegraphics[width=0.8\linewidth]{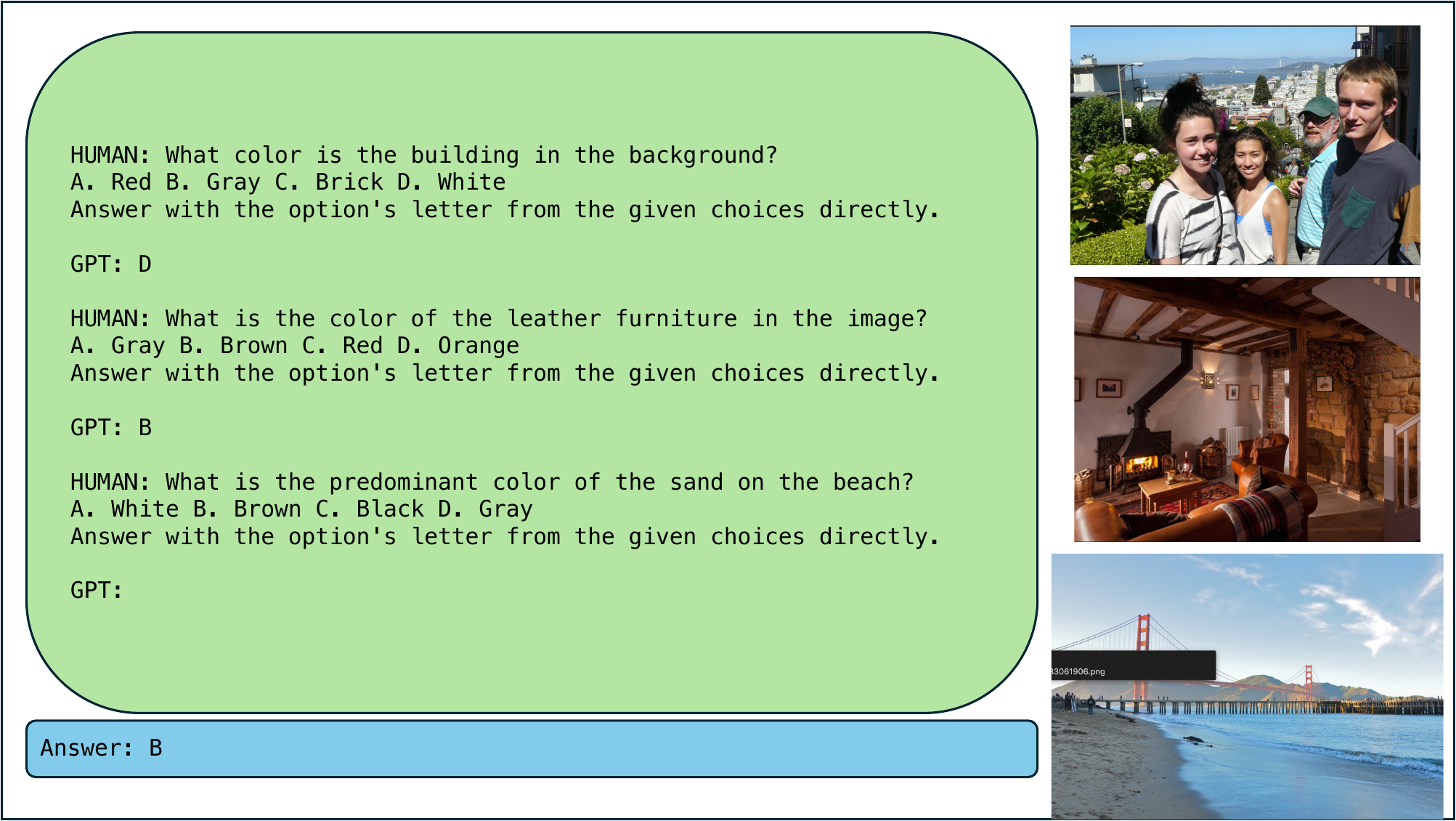}
        \caption{SEED benchmark task 3: Instance Attribute}
    \end{subfigure}
    \caption{SEED benchmark training examples tasks 1-3}
    \label{fig:seed012}
\end{figure}

\begin{figure}[h]
    \begin{subfigure}{\textwidth}
        \centering
        \includegraphics[width=0.8\linewidth]{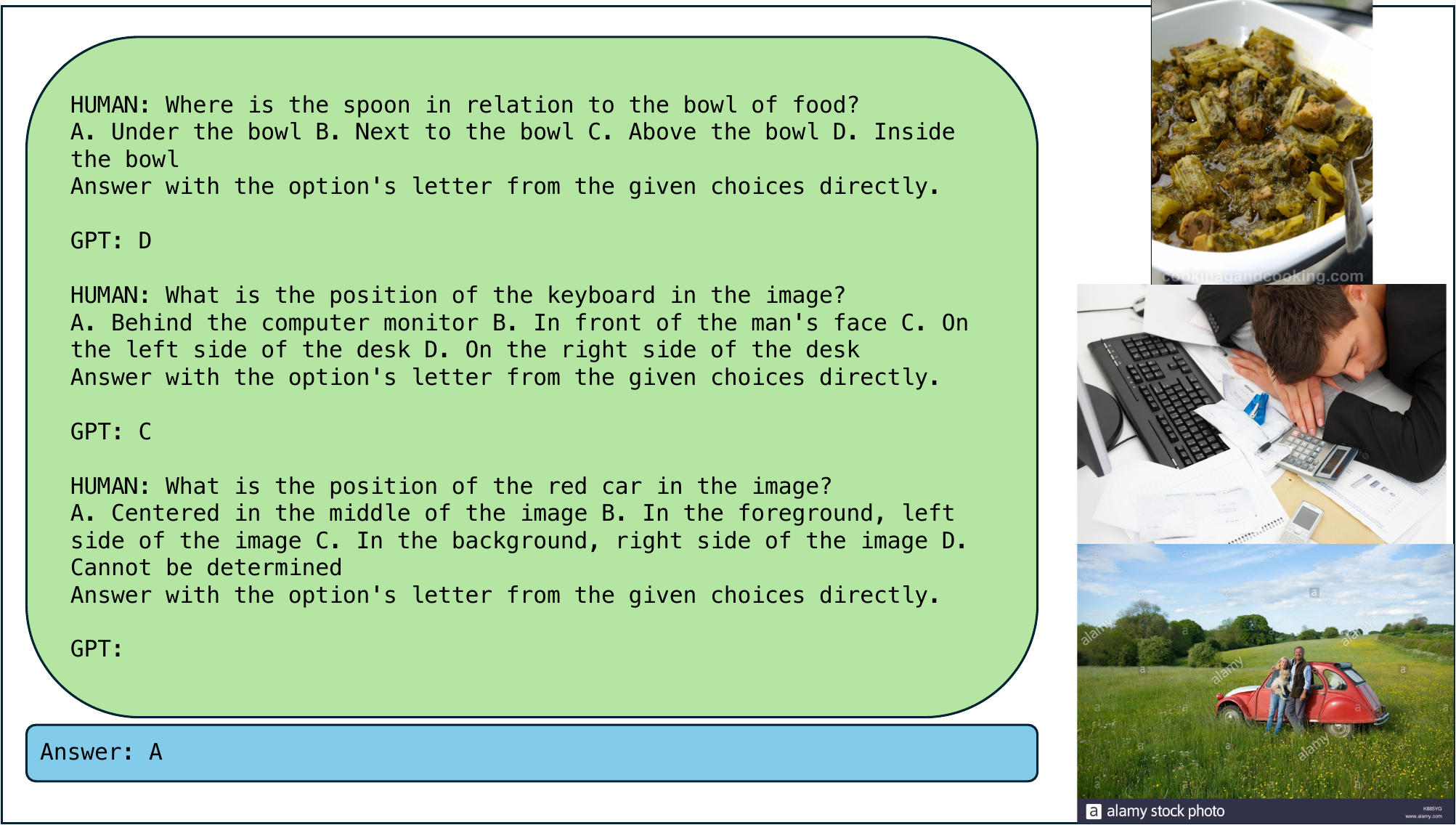}
        \subcaption{SEED benchmark task 4: Instance Location}
        \label{subfig:d}
    \end{subfigure}

    \begin{subfigure}{\textwidth}
        \centering
        \includegraphics[width=0.8\linewidth]{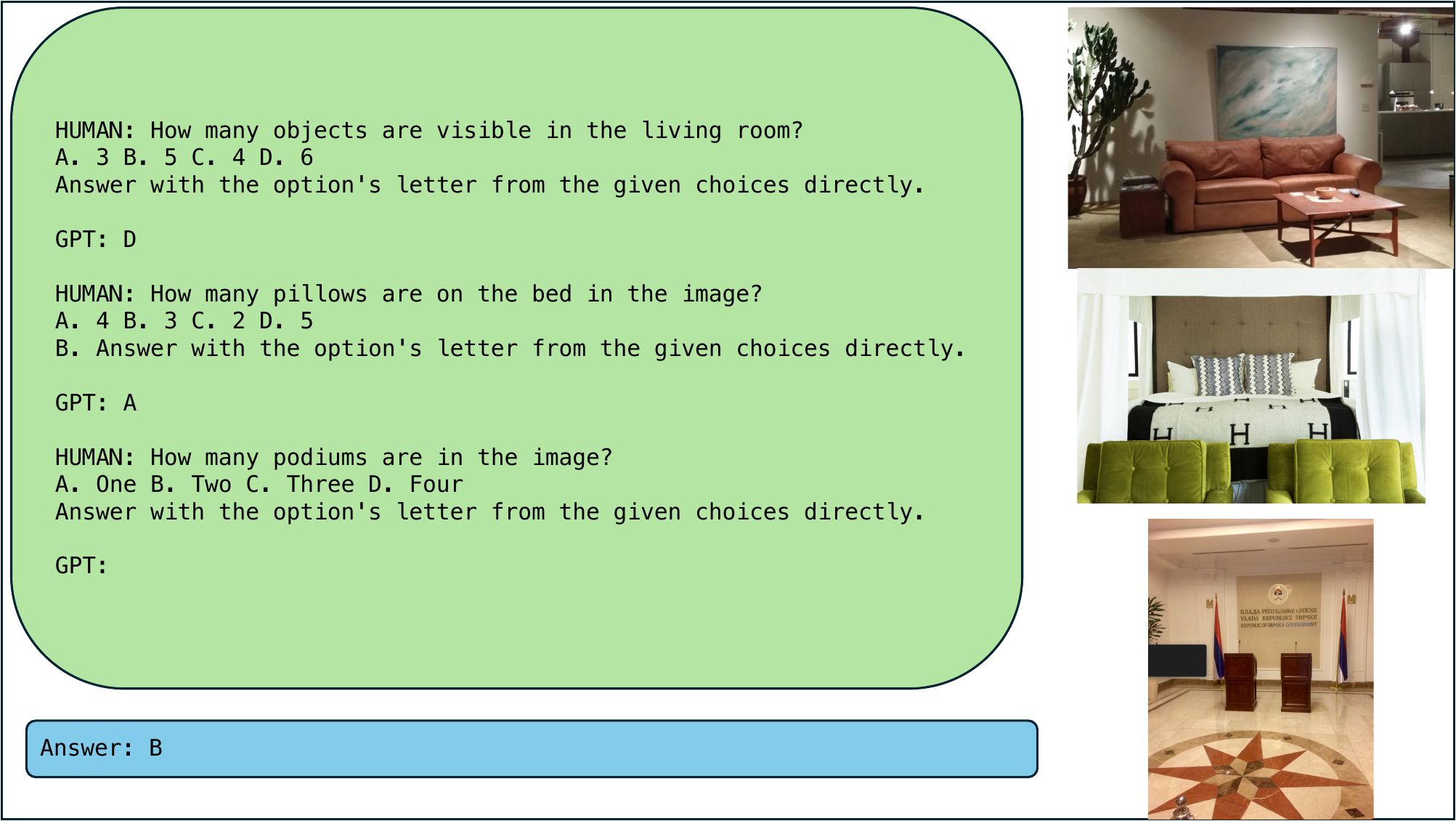}
        \subcaption{SEED benchmark task 5: Instance Counting}
        \label{subfig:e}
    \end{subfigure}

    \caption{SEED benchmark training examples tasks 4-5}
    \label{fig:seed34}
\end{figure}

Data from tasks 1-5 from the SEED benchmark is used for creating training in-context learning instructions that share a common semantic concept within each ICL instruction, see \cref{fig:seed012,fig:seed34} for qualitative examples. 
Furthermore, we use task 1-4 data only for ICL training purposes while keeping a portion of task 5 data for ICL evaluation on Instance Counting (IC). Additionally, we generate ICL tasks from SEED benchmark tasks 6-8 for testing only (to test generalization to unseen ICL tasks) as explained in \cref{sec:test_seed}. In \cref{table:seed_train_amount} we provide the detailed statistics of the SEED-bench train data used in our train split.

\subsection{VL-Checklist}
Data from tasks 0-12 from the VL-Checklist benchmark is used for creating training in-context learning instructions with a shared common semantic concept within each ICL instruction, see figures \cref{fig:vlcap012,fig:vlcap345} for examples. 
We split data in each task to non-overlapping train and test portions to test the resulting VLM on the ICL capability for each semantic concept (task) of the VL-Checklist (as explained in \cref{sec:vlc-test-data}). In \cref{table:vl_amount} we provide the detailed statistics of the VL-Checklist train data used in our train split.

\section{ICL benchmarks- results per task}
\label{sec:icl_detailed_results}

\subsection{VL Checklist}
In \cref{table:1,table:2,table:3}, we provide results per task on the test splits of the VL-Checklist ICL tasks. We separately provide results for different ICL task types: QA-question answering in \cref{table:1}, MC - multiple choice questions in \cref{table:2}, and Cap - captioning in \cref{table:3}.

\subsection{SEED benchmark}
In \cref{table:4}, we provide results per task on the test partitions of the SEED benchmark-based ICL tasks we created.

\section{ICL instruction task types in test} 
\label{ICL_test}
\begin{figure}[h]
    \begin{subfigure}{\textwidth}
        \centering
        \includegraphics[width=0.8\linewidth]{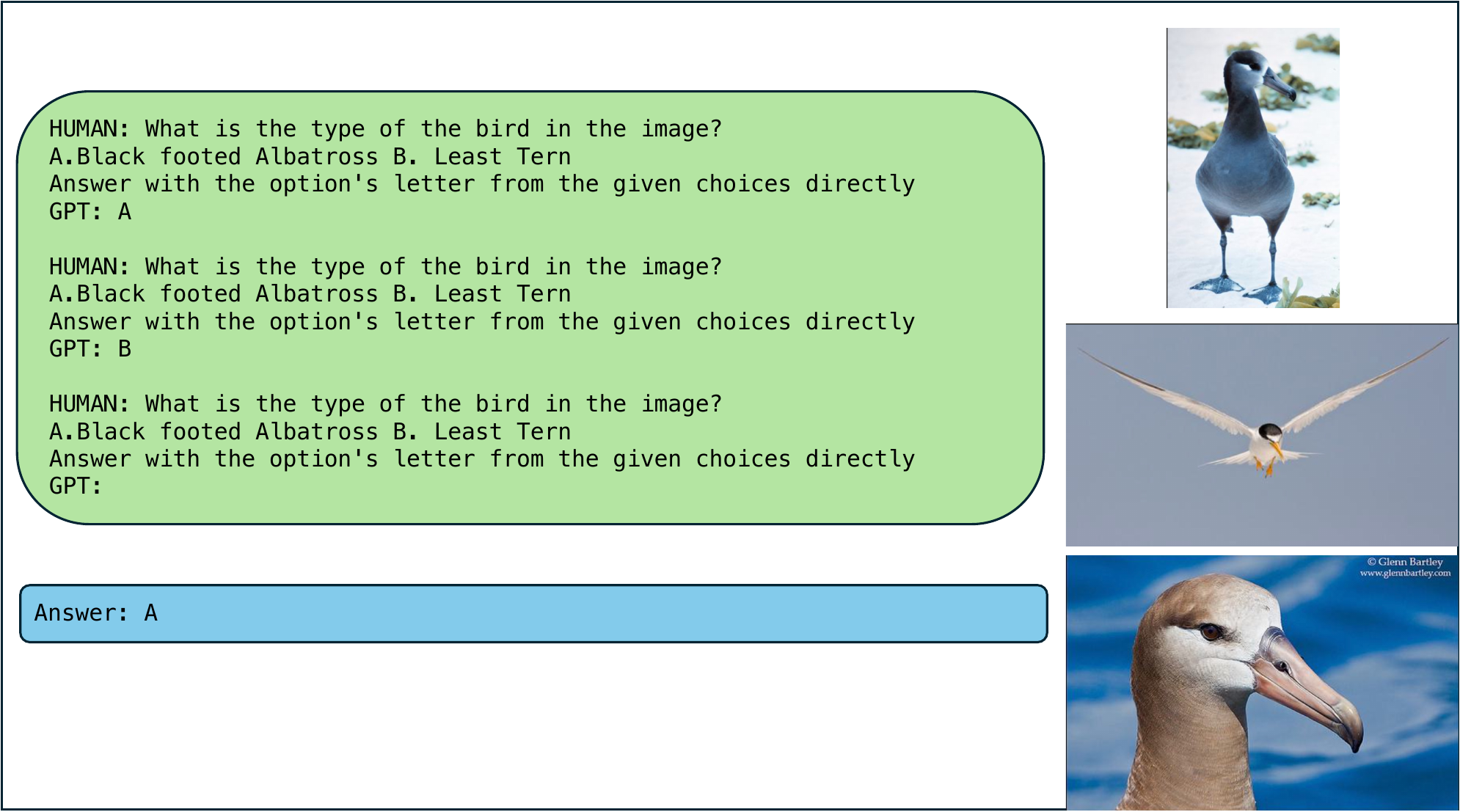}
        \caption{Few shot CUB}
    \end{subfigure}
    
    \begin{subfigure}{\textwidth}
        \centering
        \includegraphics[width=0.8\linewidth]{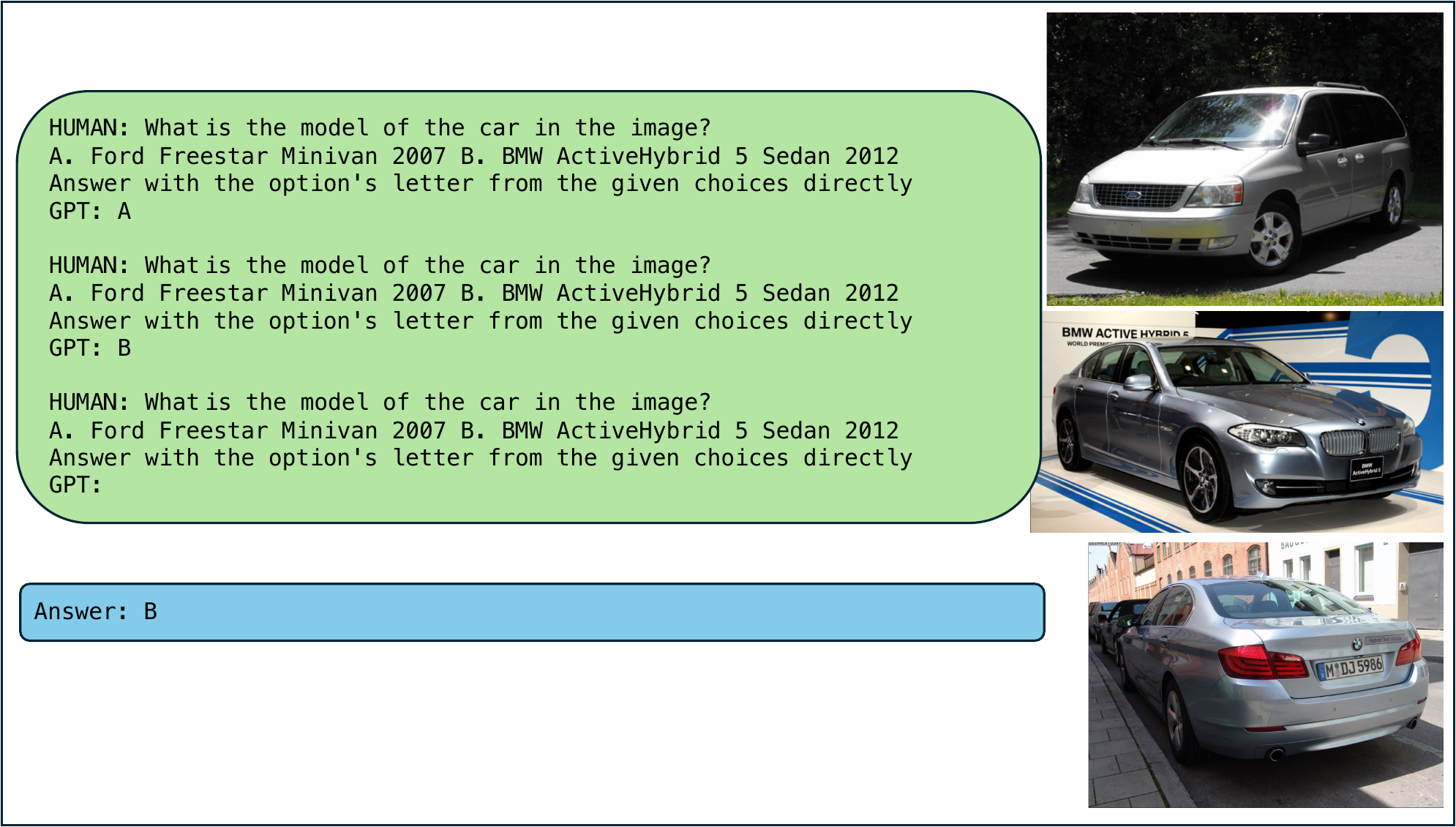}
        \caption{Few shot Stanford Cars}
    \end{subfigure}
    
    \begin{subfigure}{\textwidth}
        \centering
        \includegraphics[width=0.8\linewidth]{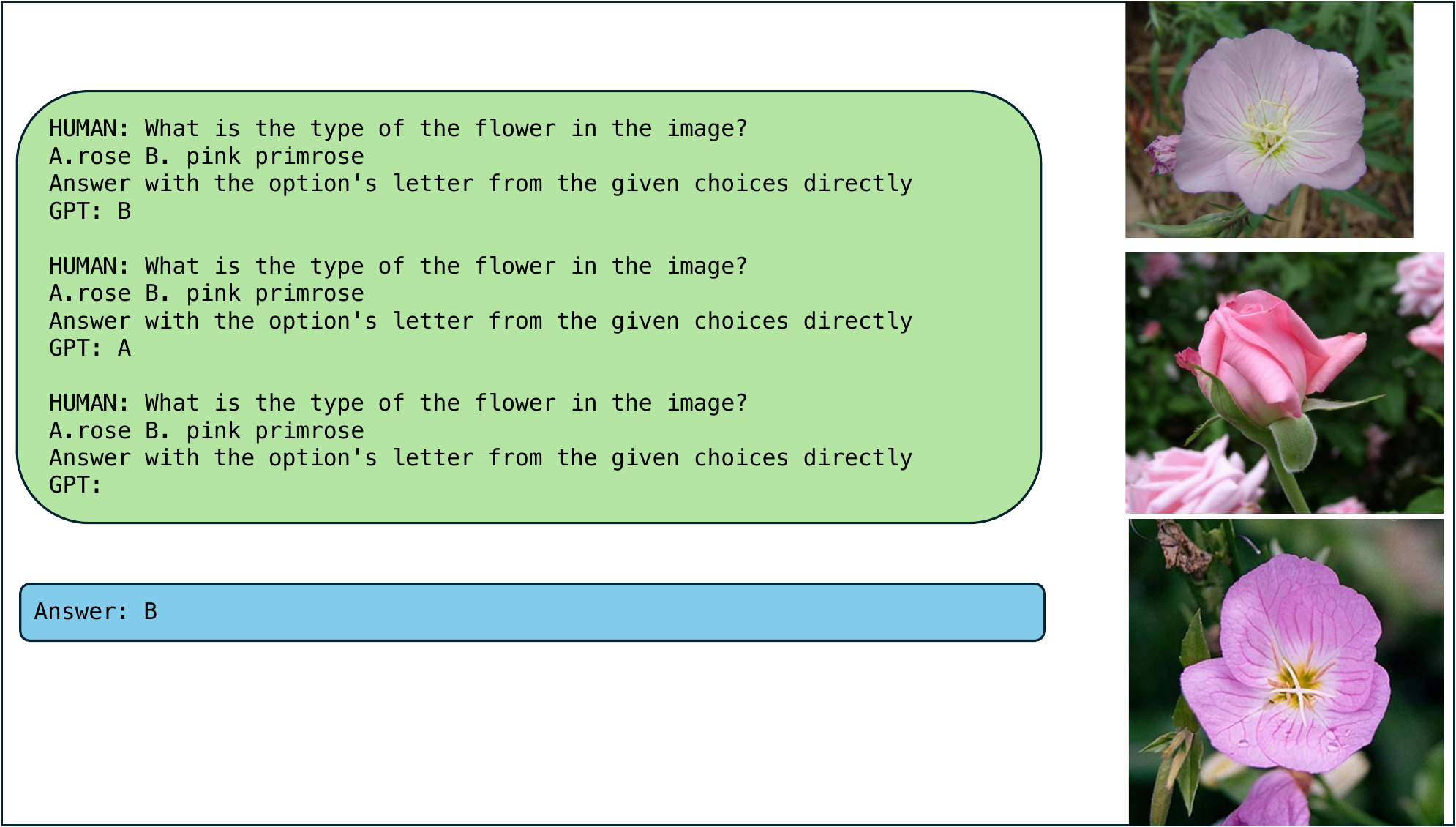}
        \caption{Few shot Flowes}
    \end{subfigure}
    
    \caption{Few shot datasets}
    \label{fig:fs1}

\end{figure}

\begin{figure}[h]
    \begin{subfigure}{\textwidth}
        \centering
        \includegraphics[width=0.8\linewidth]{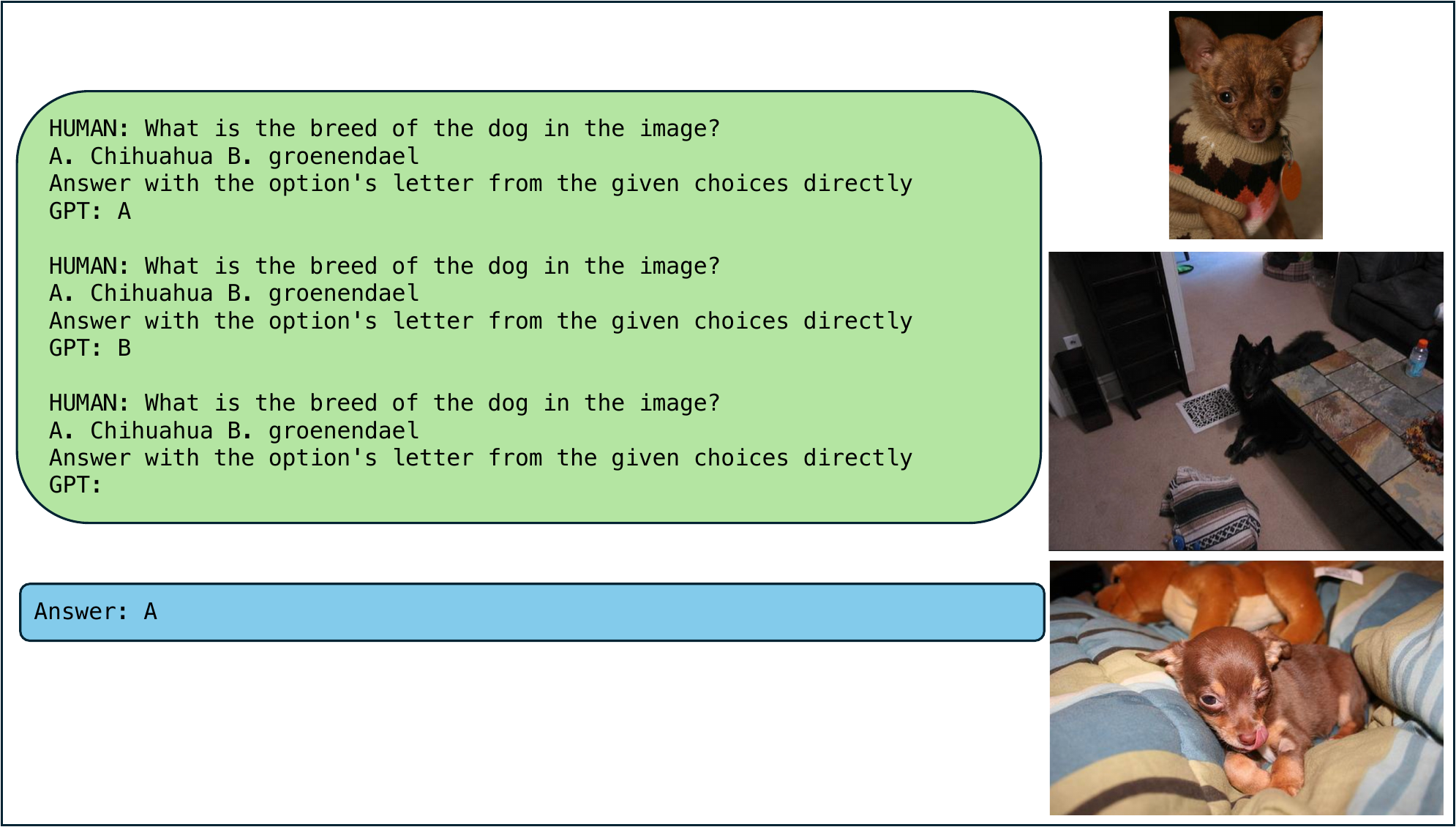}
        \caption{Few shot Stanford Dogs}
    \end{subfigure}
    
    \begin{subfigure}{\textwidth}
        \centering
        \includegraphics[width=0.8\linewidth]{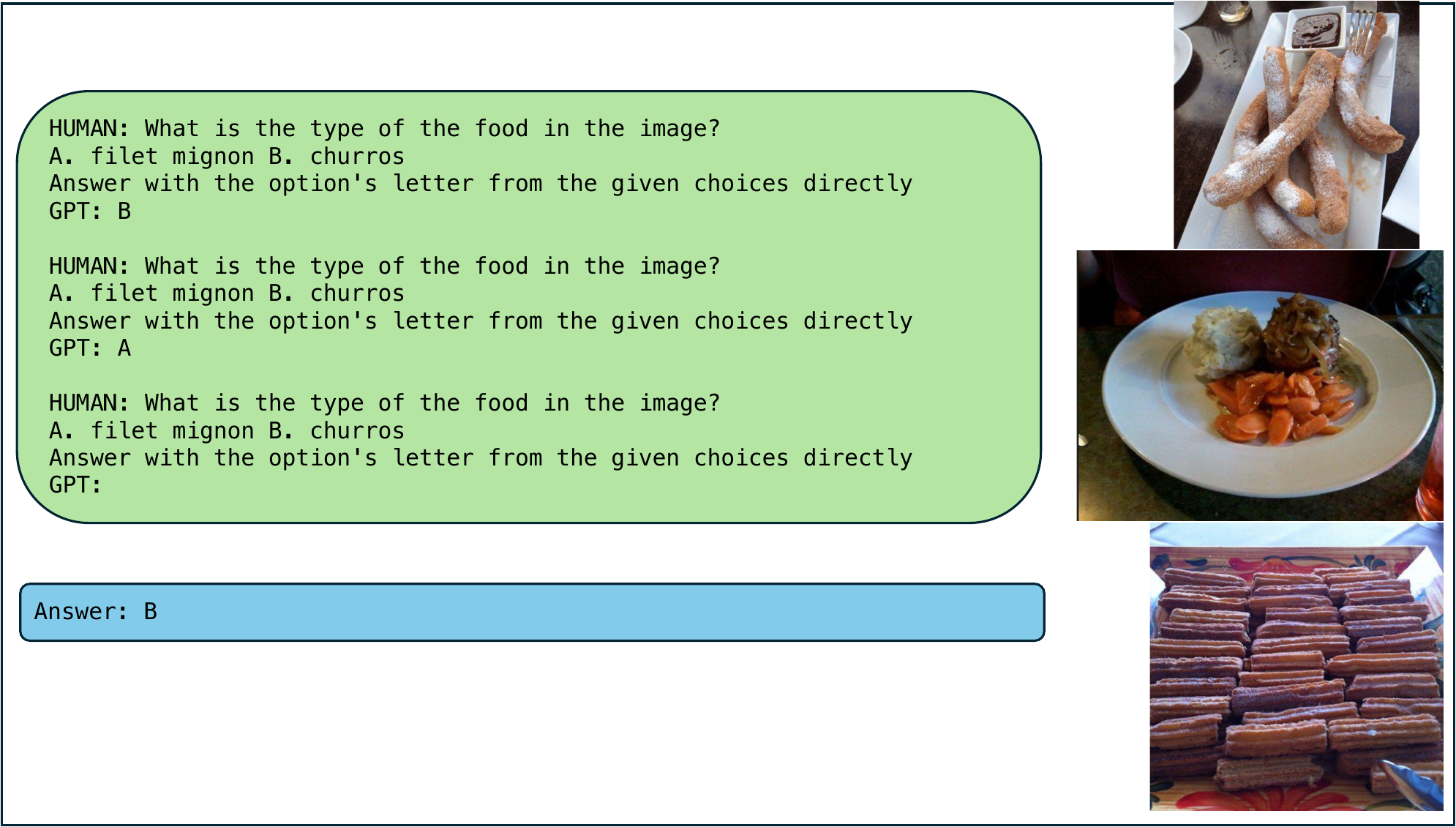}
        \caption{Few shot Food101}
    \end{subfigure}
    
    \caption{few shot datasets}
    \label{fig:fs2}

\end{figure}

We show in \cref{fig:fs1,fig:fs2,fig:seed678} examples from each dataset and task used in the test set.

\subsection{Few-shot benchmarks} 
We create fine-grained, few-shot visual recognition ICL benchmarks from the following fine-grained classification datasets: Stanford Dogs~\cite{dogs}, CUBS~\cite{cubs200}, FOOD-101~\cite{food}, Stanford Cars~\cite{cars}, Flowers~\cite{flowers}.

\subsection{SEED benchmark tests examples}\label{sec:test_seed}
\begin{figure}[h]
    \begin{subfigure}{\textwidth}
        \centering
        \includegraphics[width=0.8\linewidth]{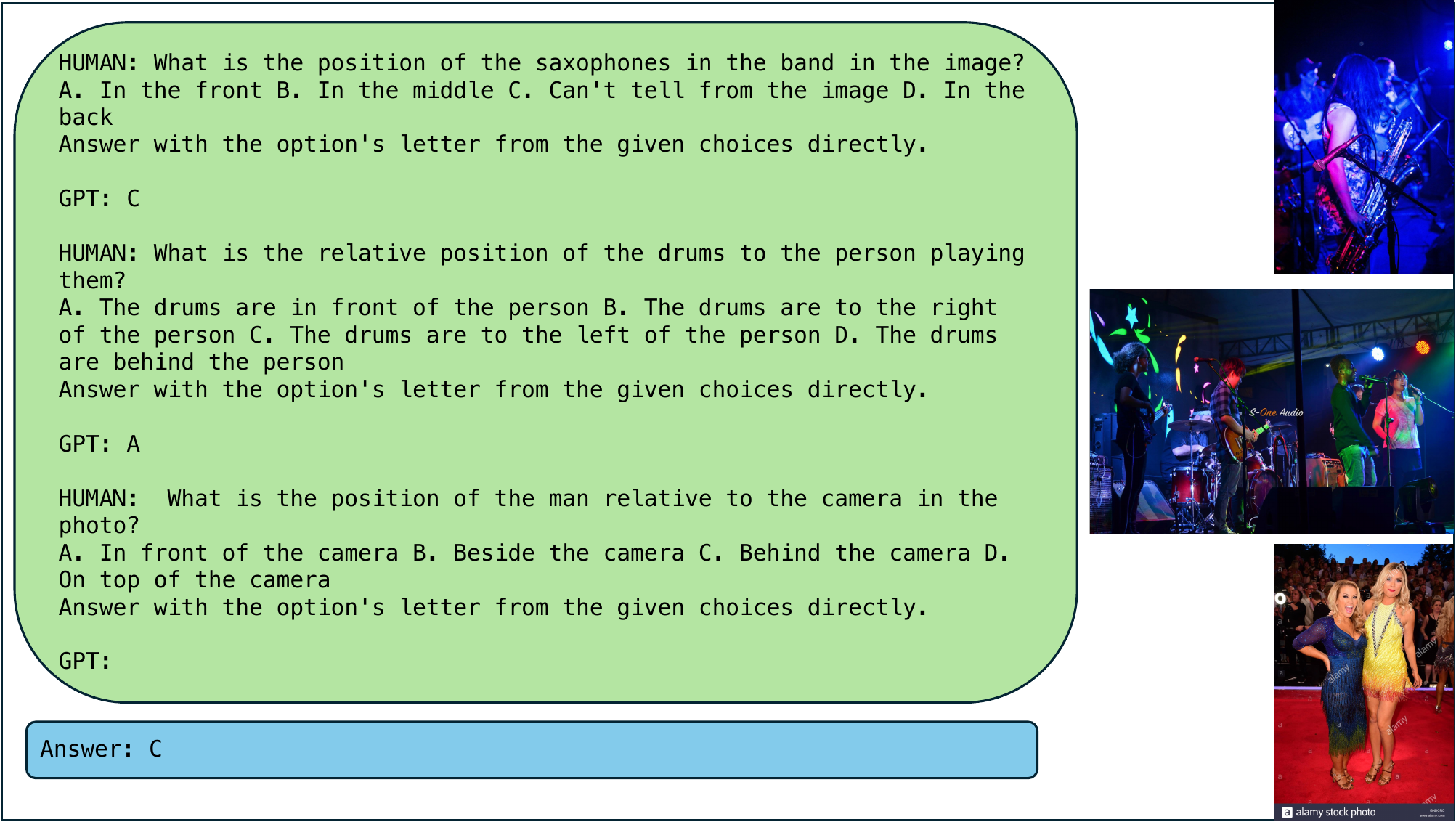}
        \caption{SEED benchmark task 6: Spatial Relation}
    \end{subfigure}
    
    \begin{subfigure}{\textwidth}
        \centering
        \includegraphics[width=0.8\linewidth]{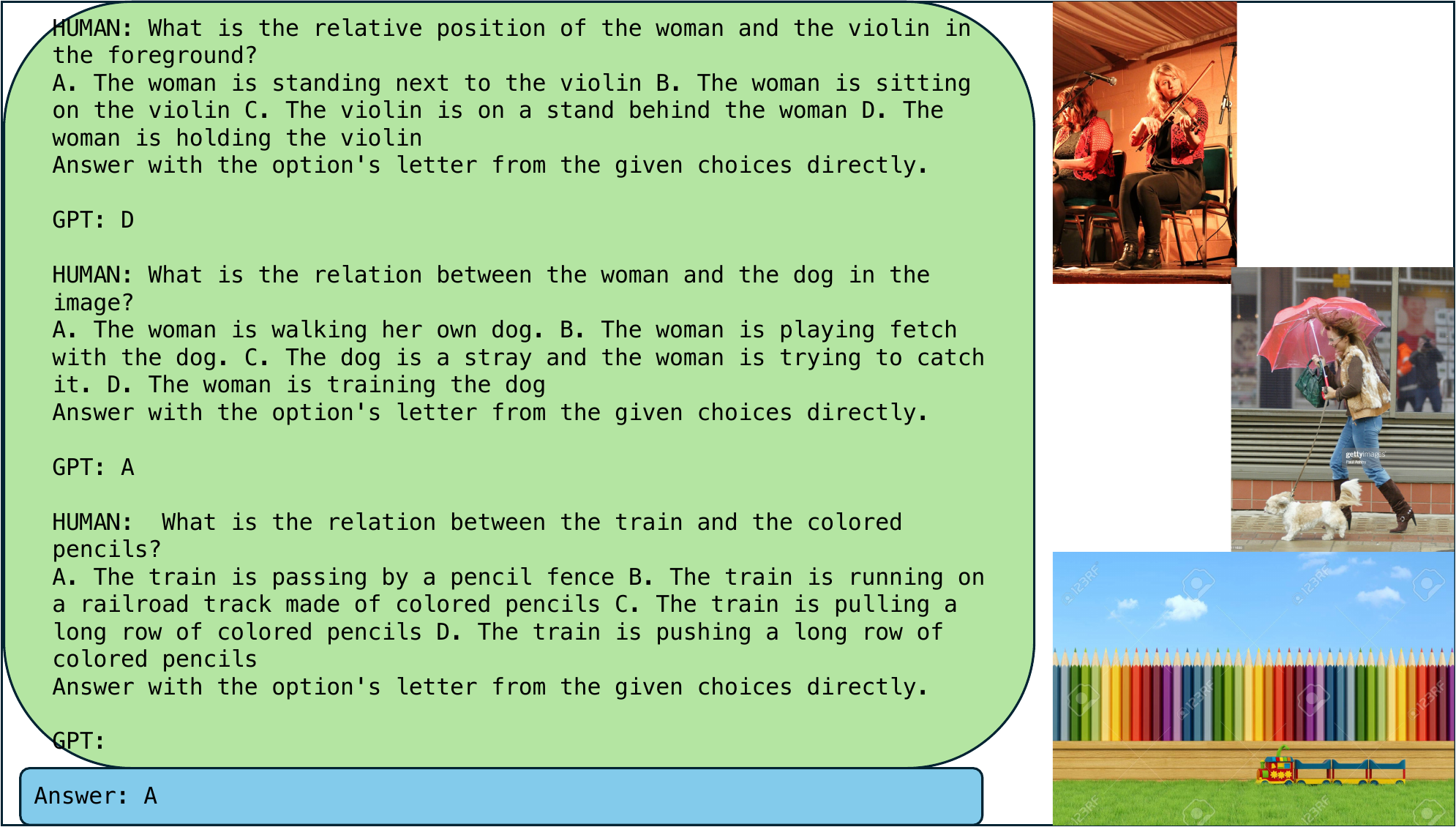}
        \caption{SEED benchmark task 7: Instance Interactions}
    \end{subfigure}
    
    \begin{subfigure}{\textwidth}
        \centering
        \includegraphics[width=0.8\linewidth]{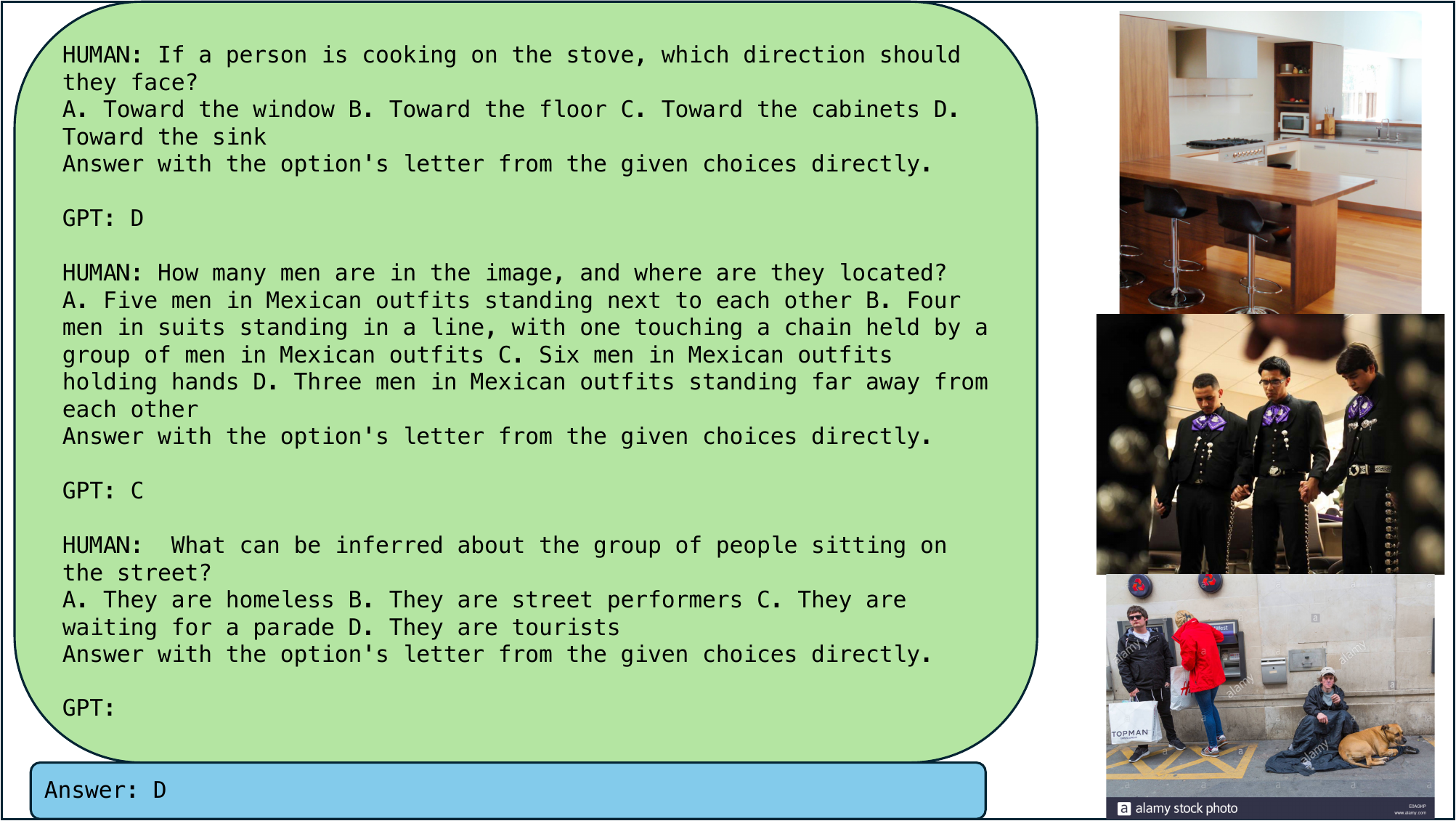}
        \caption{SEED benchmark task 8: Visual Reasoning}
    \end{subfigure}
    
    \caption{SEED benchmark Test examples tasks 6-8}
        \label{fig:seed678}

\end{figure}

For ICL testing on SEED benchmark, we use a test split for SEED benchmark task 5 (Instance Counting) that we created (`seen ICL task' - observed during training that contained a train partition of task 5), as well as the entire data of SEED benchmark tasks 6-8 (used for testing only, checking generalization of our ICL instruction tuning approach to `unseen ICL tasks' - unobserved during training). 
\cref{fig:seed678} provides examples of the unseen ICL tasks from SEED benchmark - 6 7 8. We are showing these examples already formatted as ICL tasks. In \cref{table:seed_test_amount} we provide the detailed statistics of the SEED-bench test data used in our test split.

\subsection{VL-Checklist test examples} \label{sec:vlc-test-data}
\begin{figure}[ht]
    \begin{subfigure}{\textwidth}
        \centering
        \includegraphics[width=0.8\linewidth]{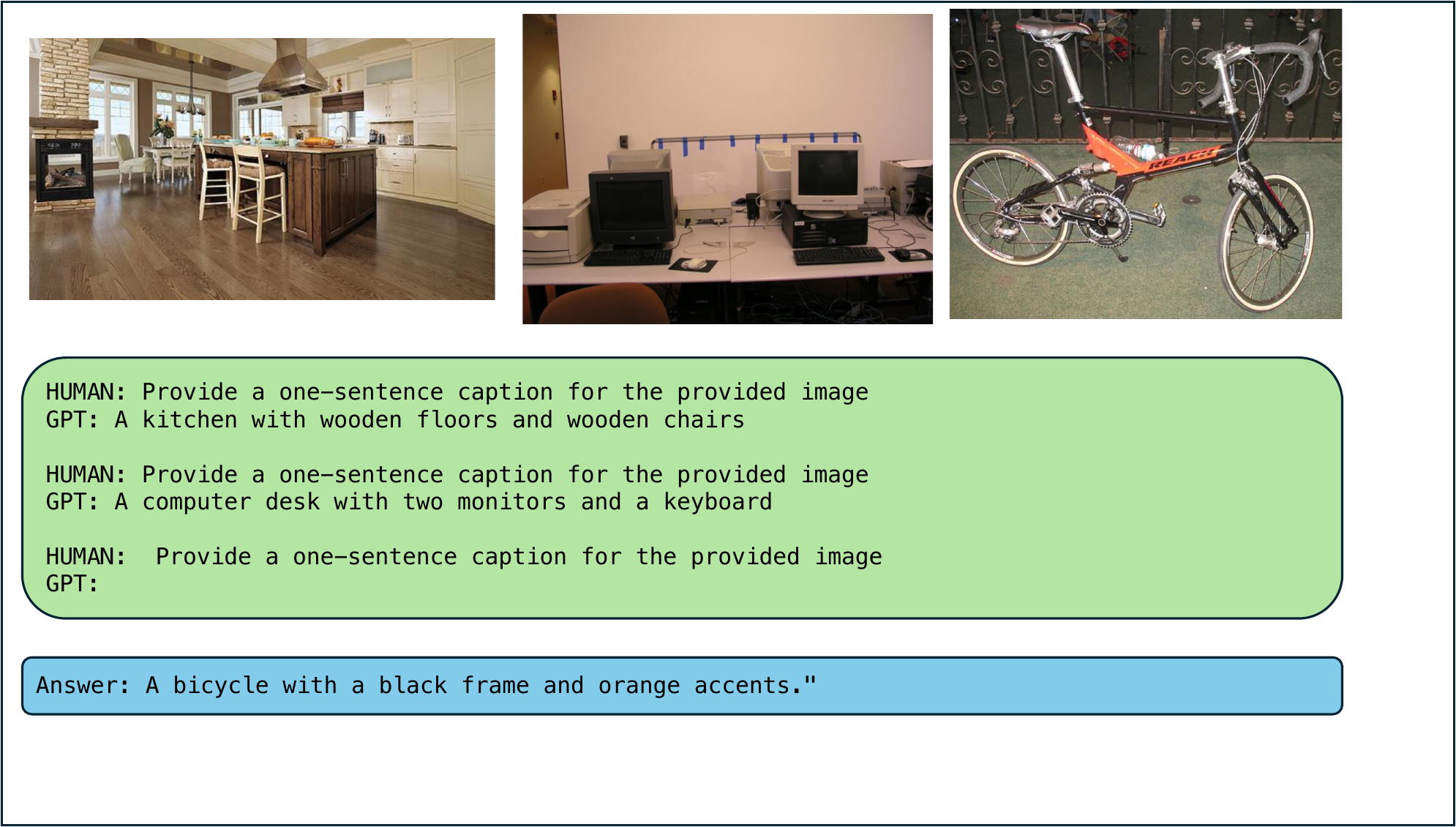}
        \caption{VL Checklist Caption task 0: Material}
    \end{subfigure}
    
    \begin{subfigure}{\textwidth}
        \centering
        \includegraphics[width=0.8\linewidth]{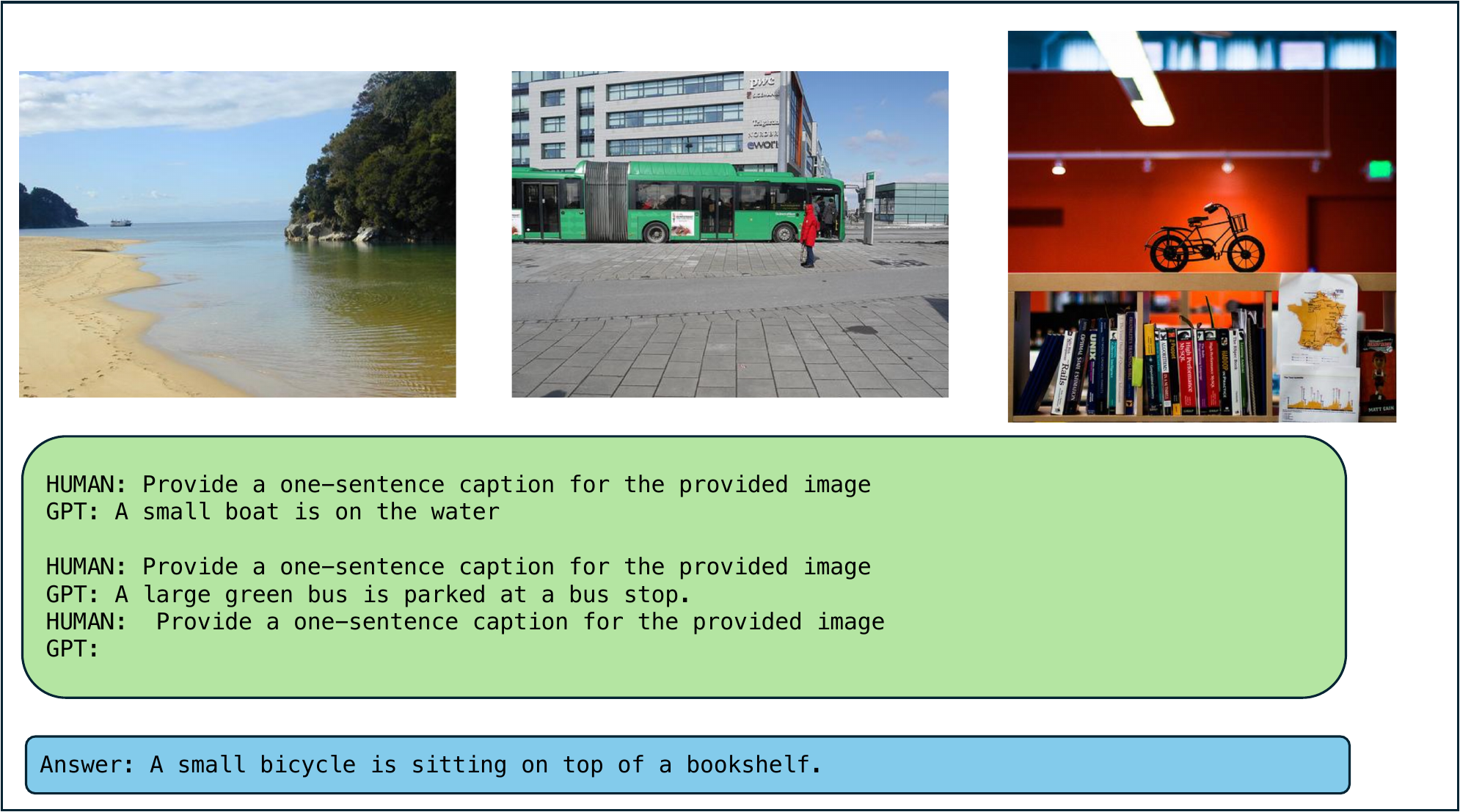}
        \caption{VL Checklist Caption task 1: Size}
    \end{subfigure}
    
    \begin{subfigure}{\textwidth}
        \centering
        \includegraphics[width=0.8\linewidth]{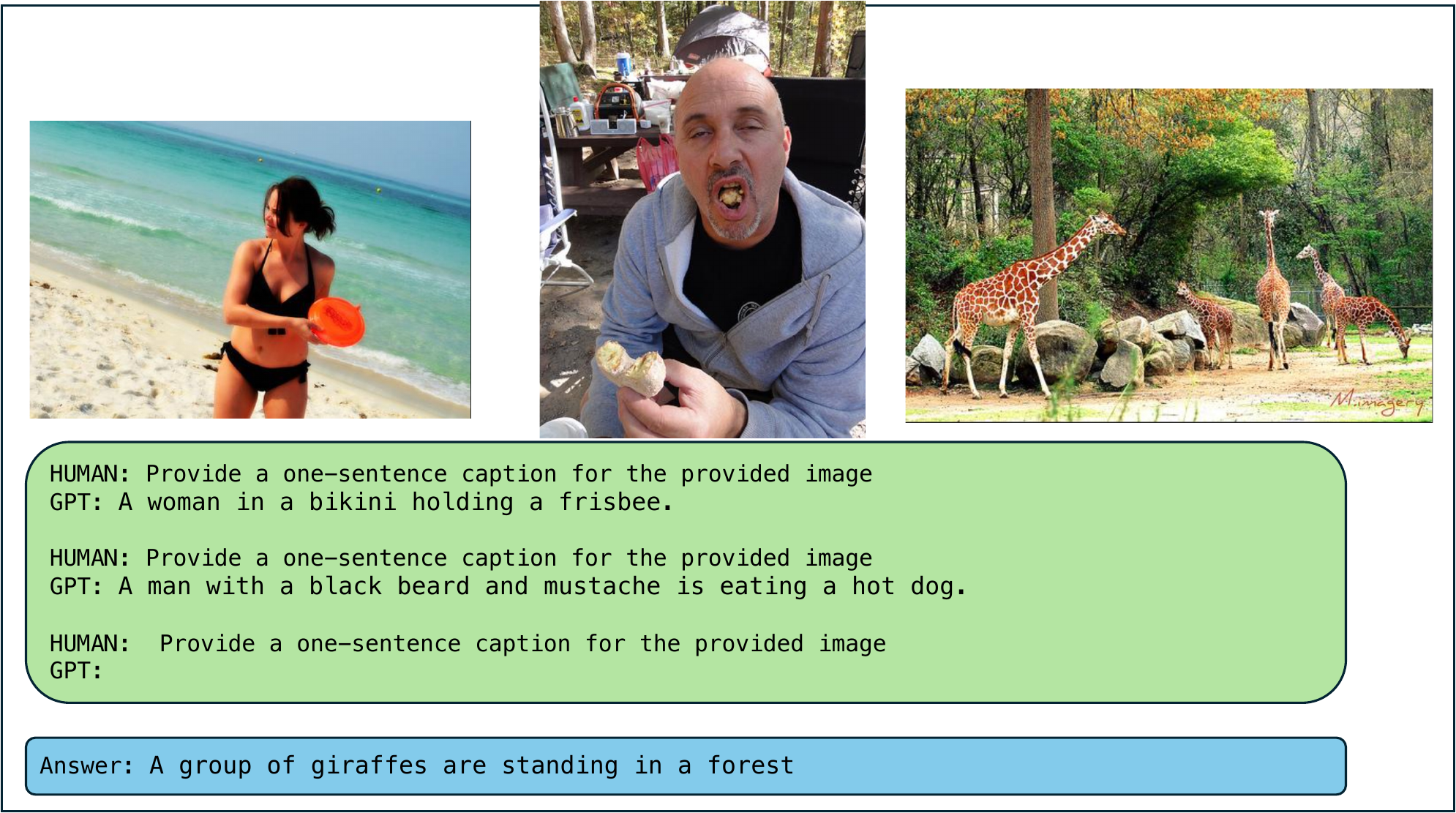}
        \caption{VL Checklist Caption task 2: Action}
    \end{subfigure}
    \caption{VL Checklist captioning training examples tasks 0-2}
    \label{fig:vlcap012}
\end{figure}

\begin{figure}[ht]
    \begin{subfigure}{\textwidth}
        \centering
        \includegraphics[width=0.8\linewidth]{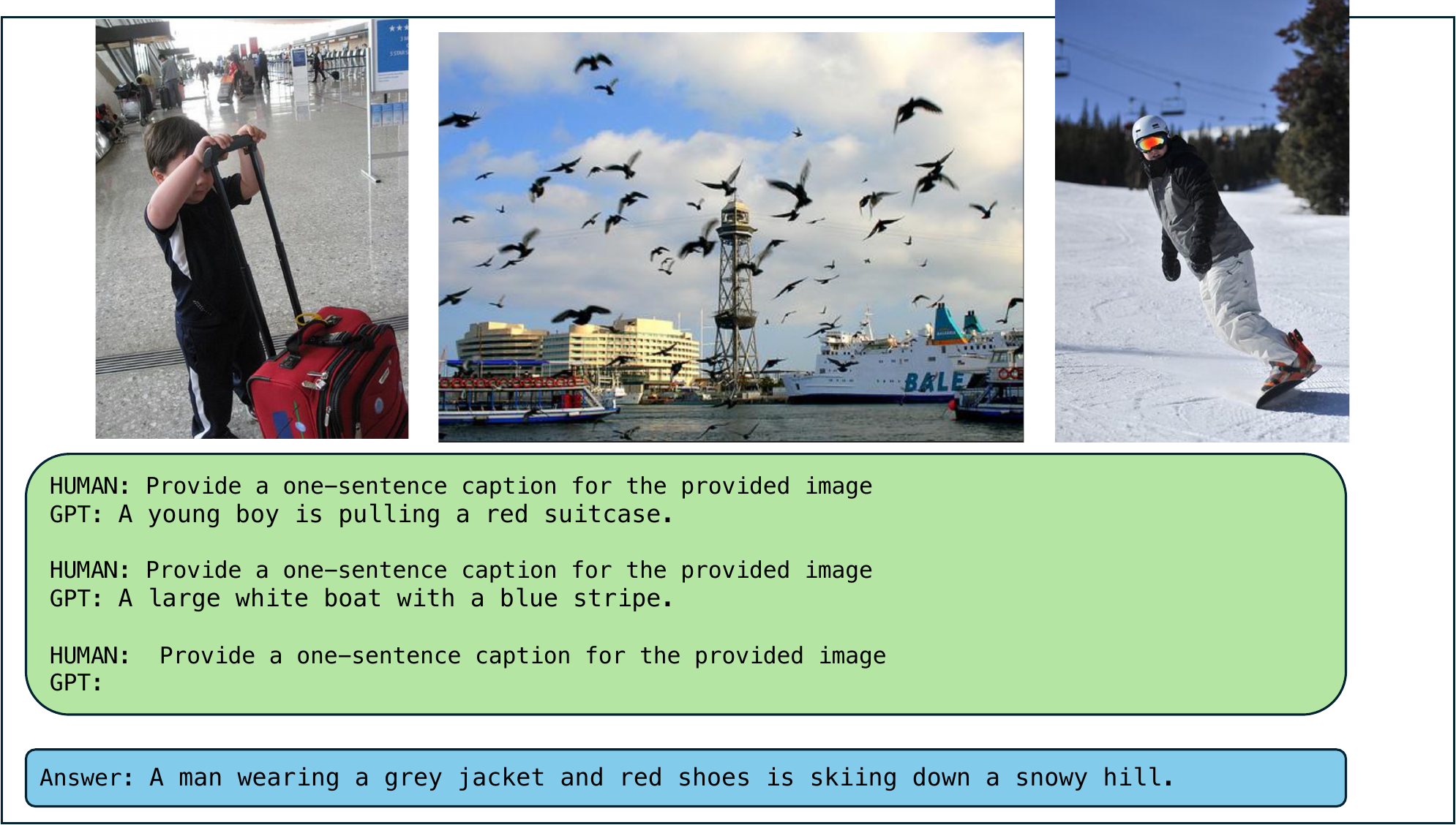}
        \caption{VL Checklist Caption task 3: Color}
    \end{subfigure}
    
    \begin{subfigure}{\textwidth}
        \centering
        \includegraphics[width=0.8\linewidth]{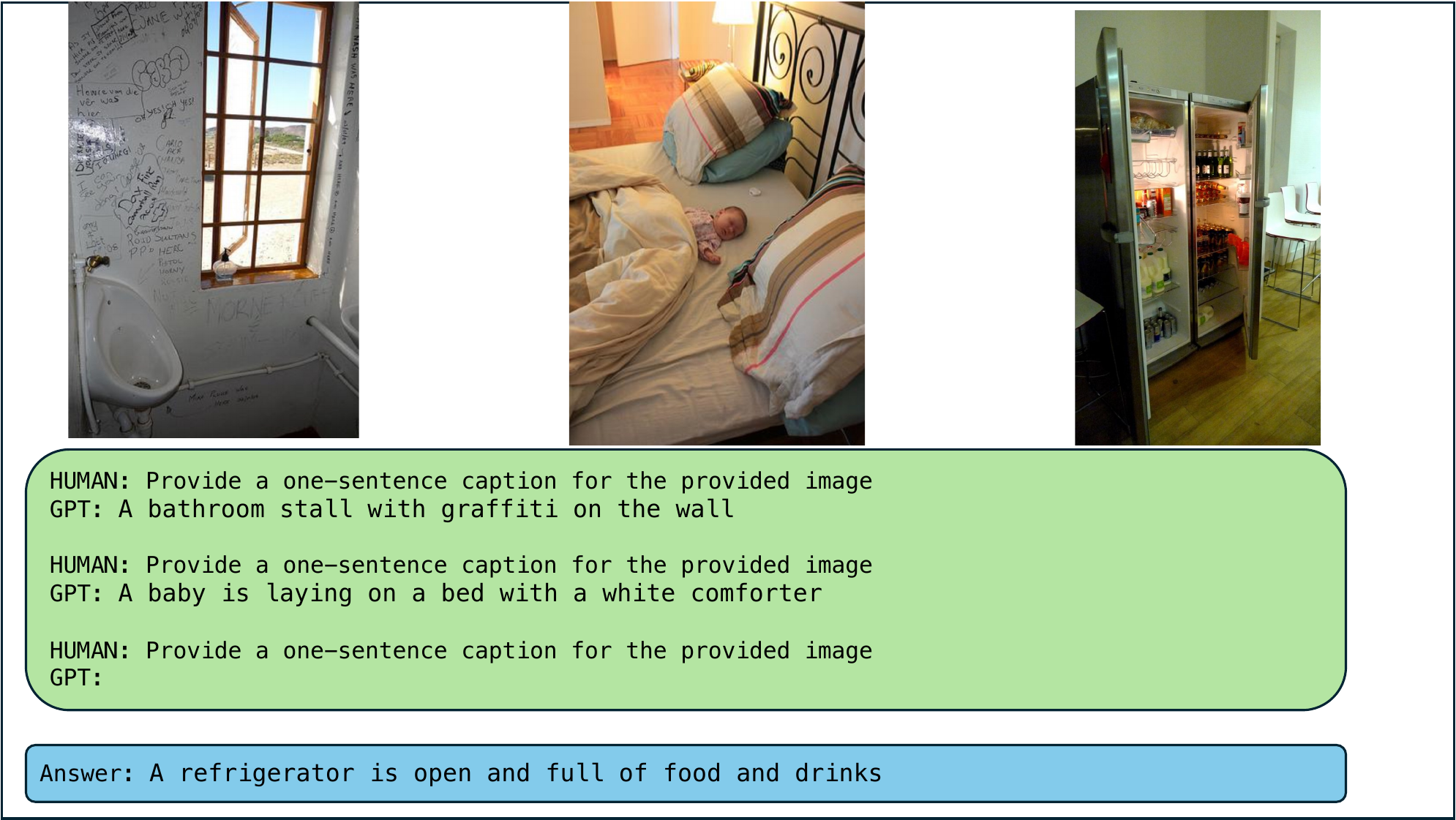}
        \caption{VL Checklist Caption task 4: State}
    \end{subfigure}
    
    \begin{subfigure}{\textwidth}
        \centering
        \includegraphics[width=0.8\linewidth]{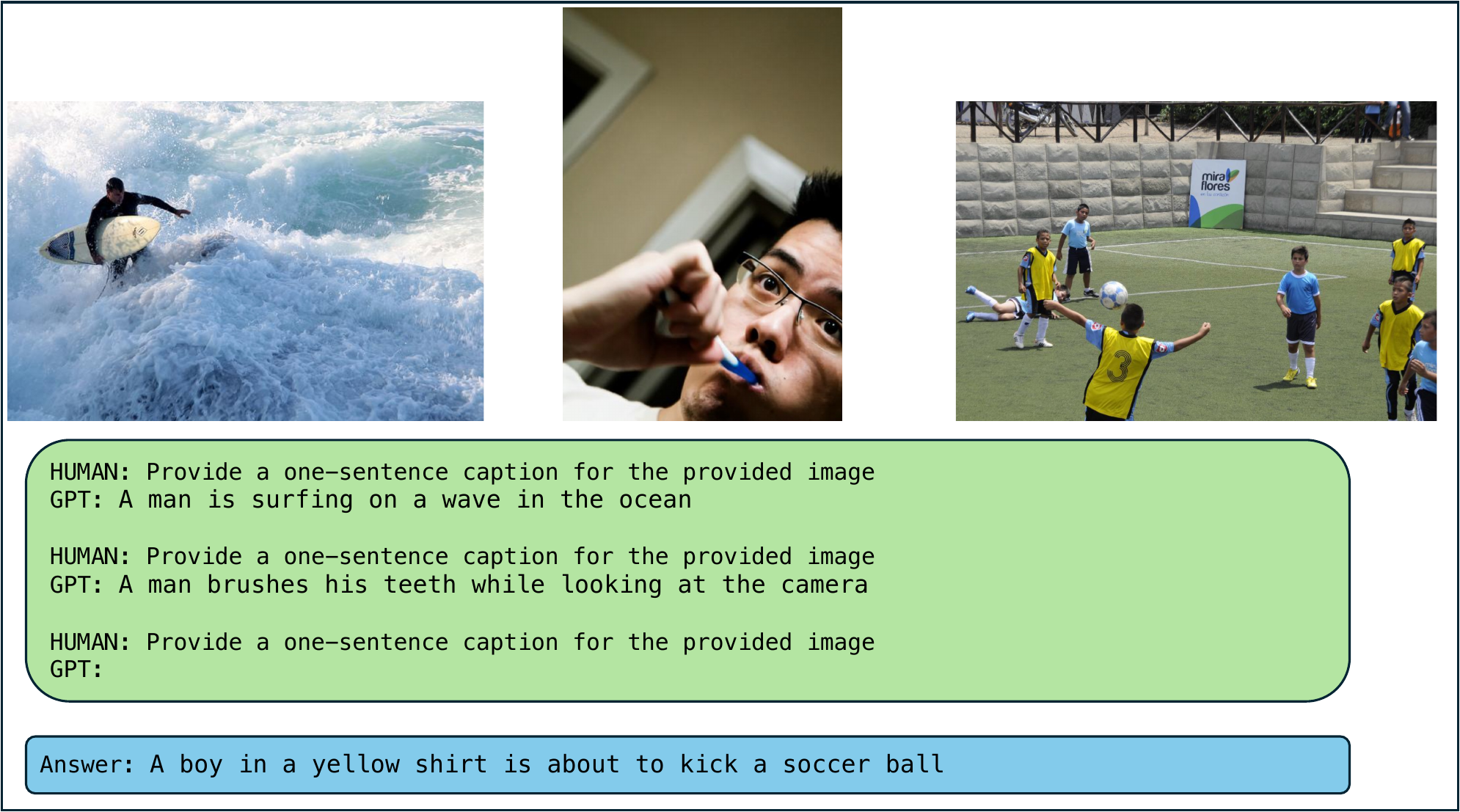}
        \caption{VL Checklist Caption task 5: Relative Action}
    \end{subfigure}
    \caption{VL Checklist captioning training examples tasks 3-5}
    \label{fig:vlcap345}
\end{figure}

For our study, we utilized the dataset provided by the VL-Checklist paper. We meticulously divided it into training and test sets through a random selection process, ensuring that there was no overlap of images between the two sets. We provide examples in the train-set VL checklist \cref{fig:vlcap012,fig:vlcap345}.
For this reason, we will not provide examples from the test set, as it would be redundant. In \cref{table:vl_amount} we provide the detailed statistics of the VL-Cheklist test data used in our test split.

\begin{table}[h]
\centering
\begin{tabular}{l|ccccccccccccc}
\toprule
Model & material & size & action & color & state & RelA & RelS & L & S & M & center & margin & mid \\
\midrule
Ours & 0.35 & 0.36 & 0.23 & 0.46 & 0.27 & 0.12 & 0.26 & 0.13 & 0.17 & 0.23 & 0.13 & 0.19 & 0.29 \\
\bottomrule
\end{tabular}
\caption{Accuracy (\%) per Question answering task in VL Checklist. RelA - relation action, RelS-relation spatial, L-Large Object, S-Small Object, M-Medium Object.}
\label{table:1}
\end{table}

\begin{table}[h]
\centering
\begin{tabular}{l|ccccccccccccc}
\toprule
Model & material & size  & action & color & state & RelA  & RelS & L & S & M & center & margin & mid   \\ 
\midrule
Ours & 0.99     & 0.76  & 0.96   & 0.94  & 0.95  & 0.98  & 0.98 & 0.96  & 0.98  & 0.97   & 0.97   & 0.99   & 0.97  \\ 
\bottomrule
\end{tabular}
\caption{Accuracy (\%) per Multiple Choice task in VL Checklist.  RelA - relation action, RelS-relation spatial, L-Large Object, S-Small Object, M-Medium Object.}
\label{table:2}
\end{table}
\begin{table*}
\begin{minipage}{0.5\textwidth}
\centering
\resizebox{\textwidth}{!}{\begin{tabular}{l|cccccc}
\toprule
Model     & material & size  & action & color & state & RelA   \\ 
\midrule
Ours & 0.21     & 0.19  & 0.36   & 0.18  & 0.21  & 0.26   \\ 
\bottomrule
\end{tabular}}
\caption{Accuracy (\%) per captioning task in VL Checklist.}
\label{table:3}
\end{minipage}
\hfill
\begin{minipage}{0.48\textwidth}
\centering
\begin{tabular}{l|cccc}
\toprule
{Model} & {IC} & {SR} & {II} & {VR} \\ \midrule
Ours & 0.65 & 0.53 & 0.74 & 0.76 \\ 

\bottomrule
\end{tabular}
\caption{Accuracy (\%) per task in SEED Bench. IC-Instance Counting, SR- Spatial Relation, II-Instance Interaction, VR- Visual Reasoning.}
\label{table:4}
\end{minipage}
\end{table*}

\begin{table}[h]
\centering
\begin{tabular}{l|cccc}
\toprule
{Split} & {IC} & {SR} & {II} & {VR} \\ \midrule
Test & 251 & 657 & 97 & 331 \\ 
\bottomrule
\end{tabular}
\caption{Amount of items in test split on SEED Bench (we use part of task 5 and all of the data of tasks 6-8 for testing). IC-Instance Counting, SR- Spatial Relation, II-Instance Interaction, VR- Visual Reasoning. The `Test' row in the table provides the amount of test images in each task.}
\label{table:seed_test_amount}
\end{table}

\begin{table}[ht]
\centering
\begin{tabular}{l|ccccc}
\toprule
{Split} & {SU} & {II} & {IA} & {IL} &{IC}\\ \midrule
Train & 3158 & 1831 & 4649 & 978 & 2196  \\ 
\bottomrule
\end{tabular}
\caption{Amount of items in train split on SEED Bench (we use all data of tasks 1-4 and part of the data of task 5 for training). SU-Scene Understanding, II-Instance Identity, IA-Instance Attribute, IL-Instance Location, IC-Instance Counting.}
\label{table:seed_train_amount}
\end{table}

\begin{table}[ht]
\centering
\resizebox{\textwidth}{!}{\begin{tabular}{l|ccccccccccccc}
\toprule
Model & material & size & action & color & state & RelA & RelS & L & S & M & center & margin & mid \\
\midrule
Train & 12000 & 12000 & 4437 & 12000 & 5292 & 12000 & 900 & 12000 & 12000 & 12000 & 12000 & 12000 & 12000 \\
Test  & 1431 & 1802 & 494 & 7317 & 588 & 12000 & 100 & 7915 & 2805 & 2999 & 7929 & 2485 & 7901 \\ \bottomrule
\end{tabular}}
\caption{Amount of samples in train and test splits of VL-Checklist tasks.}
\label{table:vl_amount}
\end{table}

\begin{figure}[ht]
    \begin{subfigure}{\textwidth}
        \centering
        \includegraphics[width=0.8\linewidth]{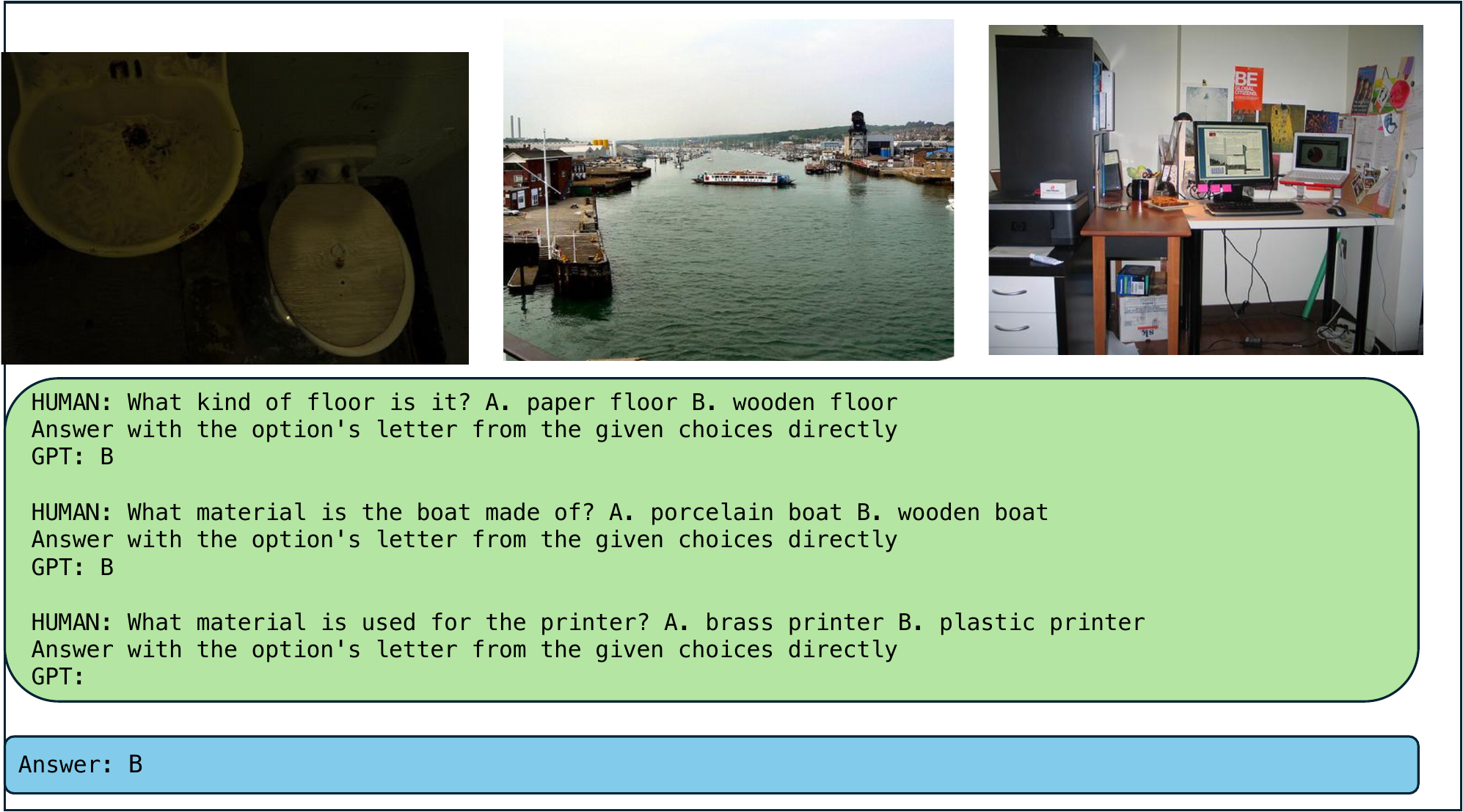}
        \caption{VL Checklist Multiple Choice task 0: Material}
    \end{subfigure}
    
    \begin{subfigure}{\textwidth}
        \centering
        \includegraphics[width=0.8\linewidth]{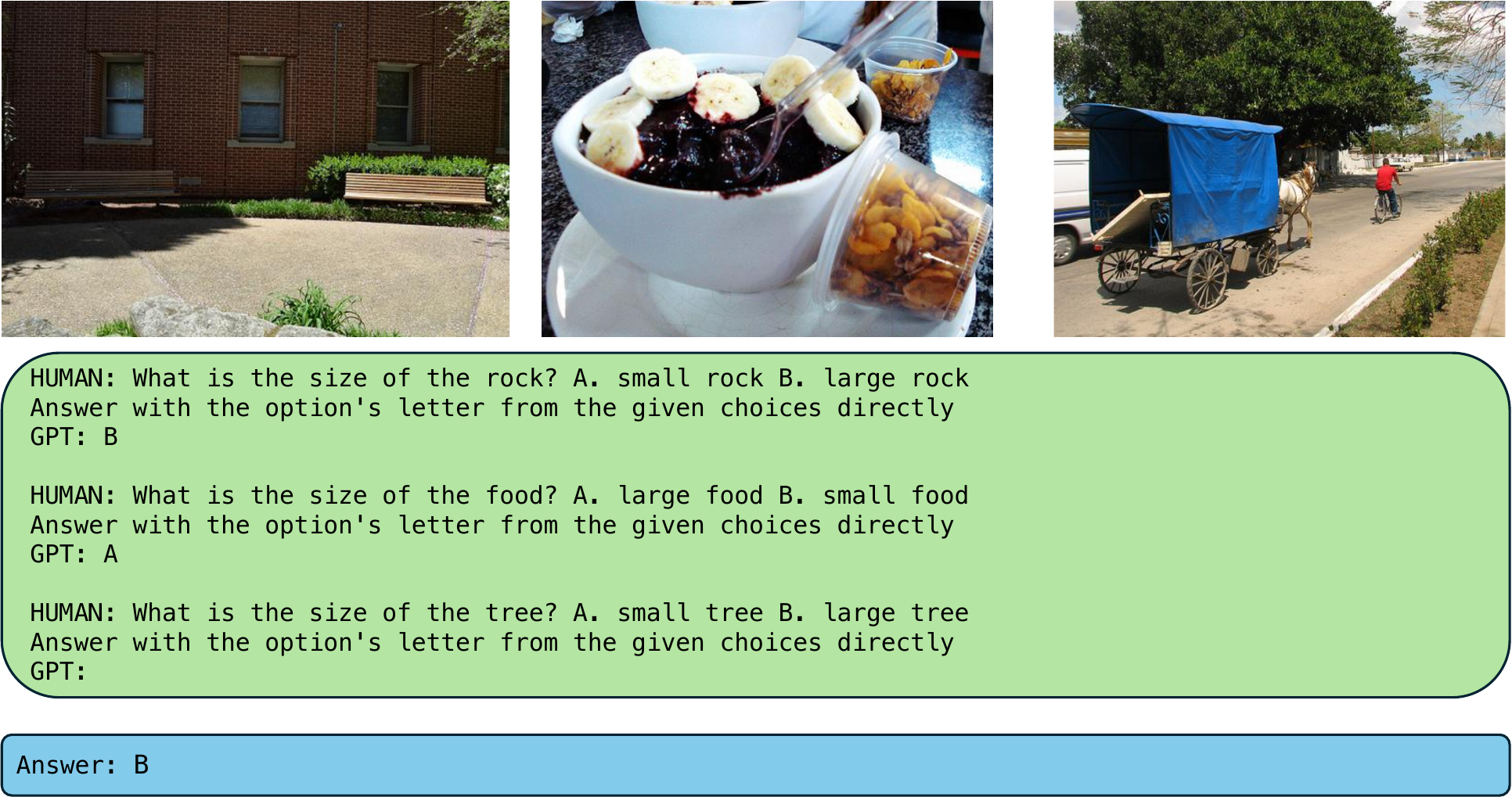}
        \caption{VL Checklist Multiple Choice task 1: Size}
    \end{subfigure}
    
    \begin{subfigure}{\textwidth}
        \centering
        \includegraphics[width=0.8\linewidth]{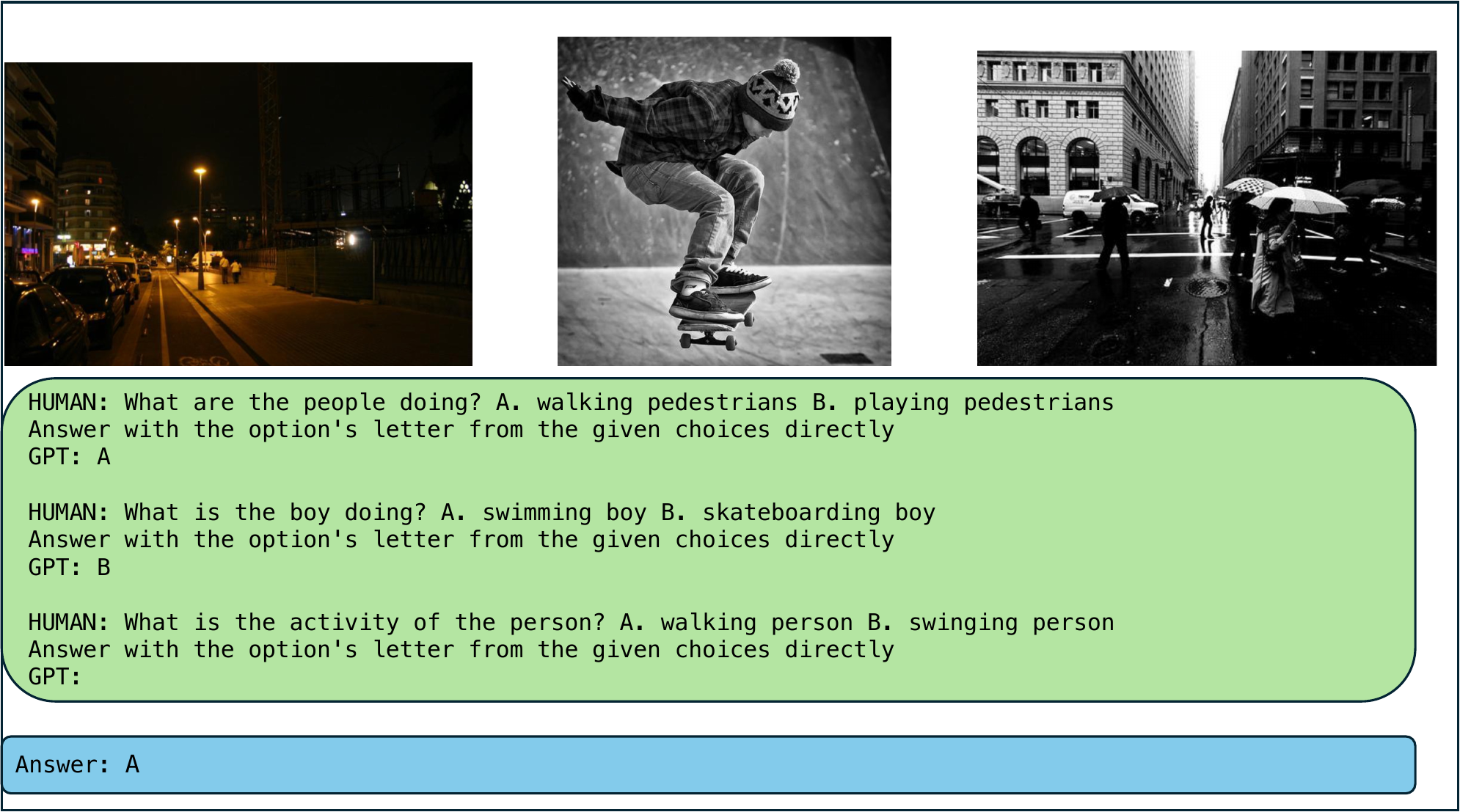}
        \caption{VL Checklist Multiple Choice task 2: action}
    \end{subfigure}
    \caption{VL Checklist Multiple Choice training examples tasks 0-2}
    \label{fig:vlmc012}

\end{figure}

\begin{figure}[h]
    \begin{subfigure}{\textwidth}
        \centering
        \includegraphics[width=0.8\linewidth]{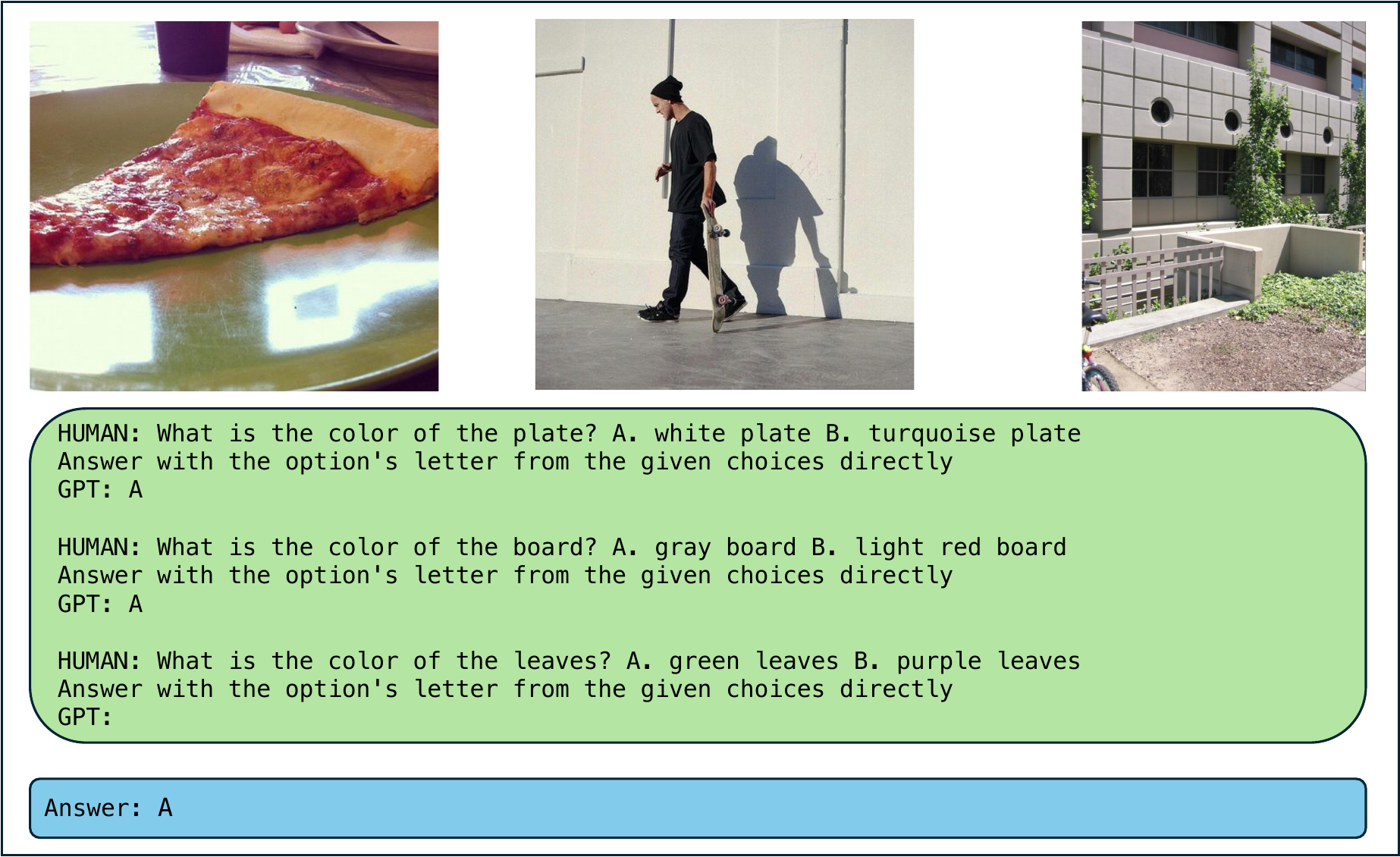}
        \caption{VL Checklist Multiple Choice task 3: Color}
    \end{subfigure}
    
    \begin{subfigure}{\textwidth}
        \centering
        \includegraphics[width=0.8\linewidth]{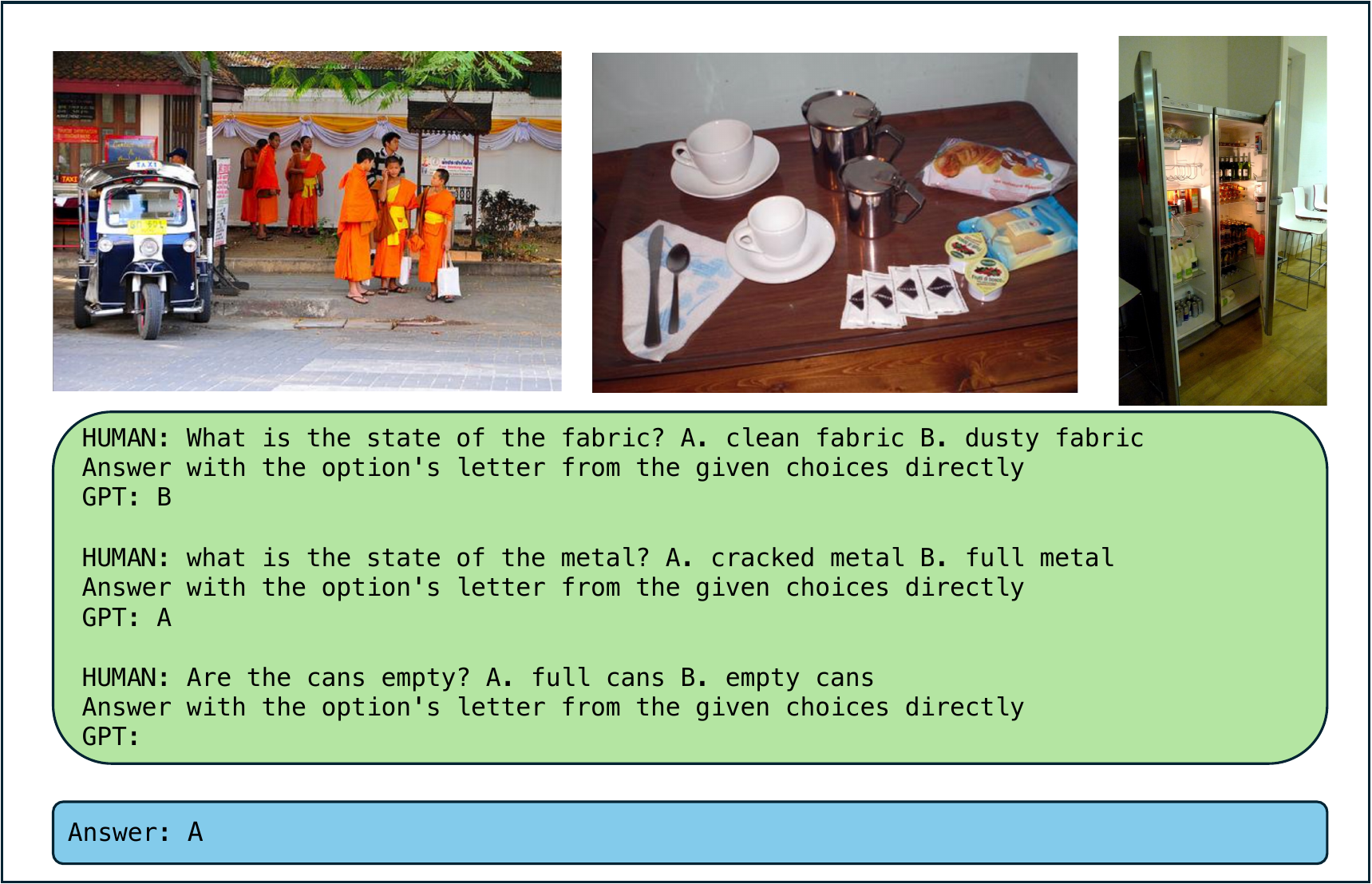}
        \caption{VL Checklist Multiple Choice task 4: State}
    \end{subfigure}
    
    \begin{subfigure}{\textwidth}
        \centering
        \includegraphics[width=0.8\linewidth]{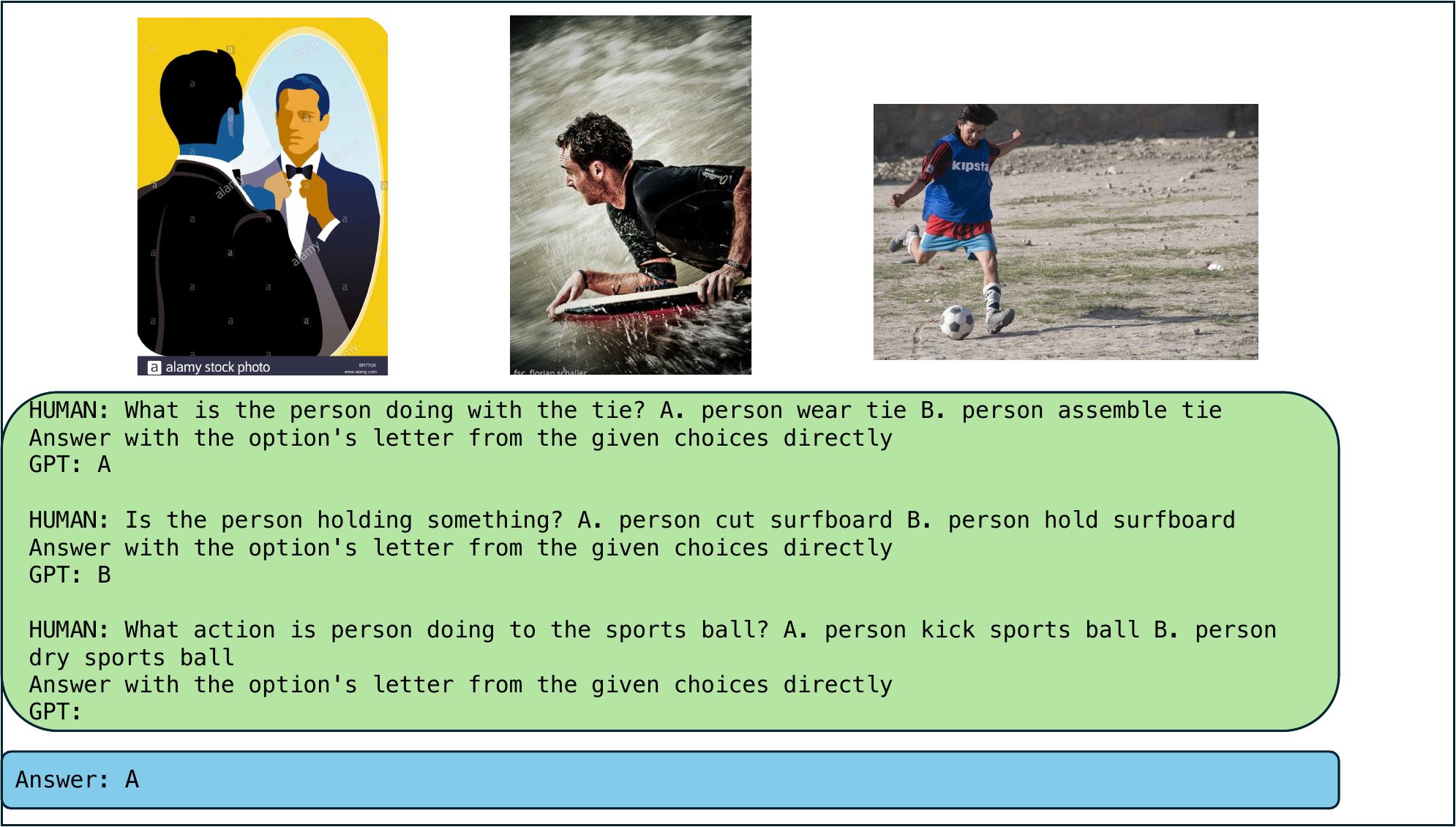}
        \caption{VL Checklist Multiple Choice task 5: Relative Action}
    \end{subfigure}
    \caption{VL Checklist Multiple Choice training examples tasks 3-5}
        \label{fig:vlmc345}

\end{figure}

\begin{figure}[h]
    \centering
    \includegraphics[width=0.9\textwidth]{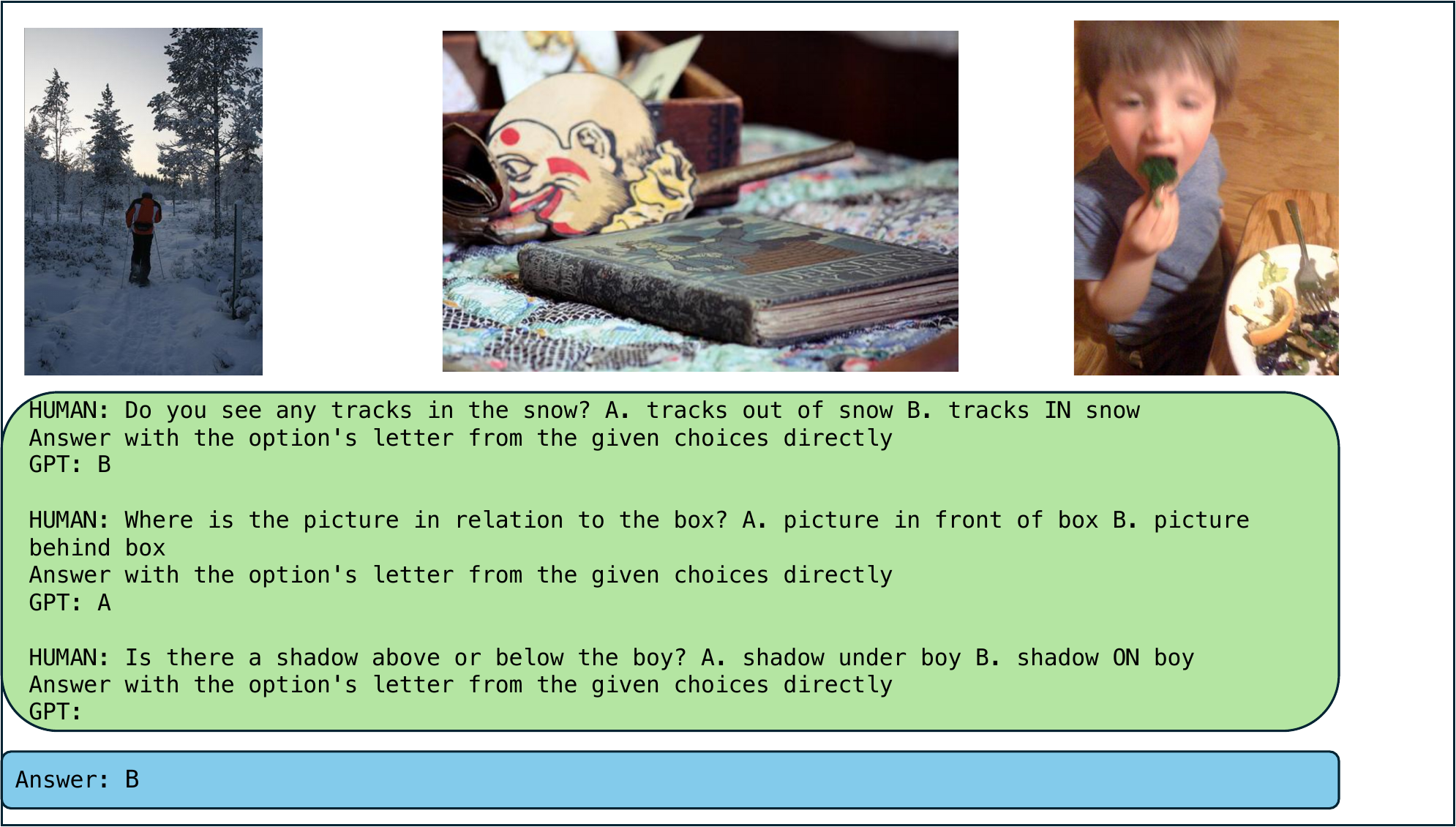}
    \caption{VL Checklist Multiple Choice task 6: Relative Spatial}
    \label{fig:vlmc6}
\end{figure}

\begin{figure}[h]
    \begin{subfigure}{\textwidth}
        \centering
        \includegraphics[width=0.8\linewidth]{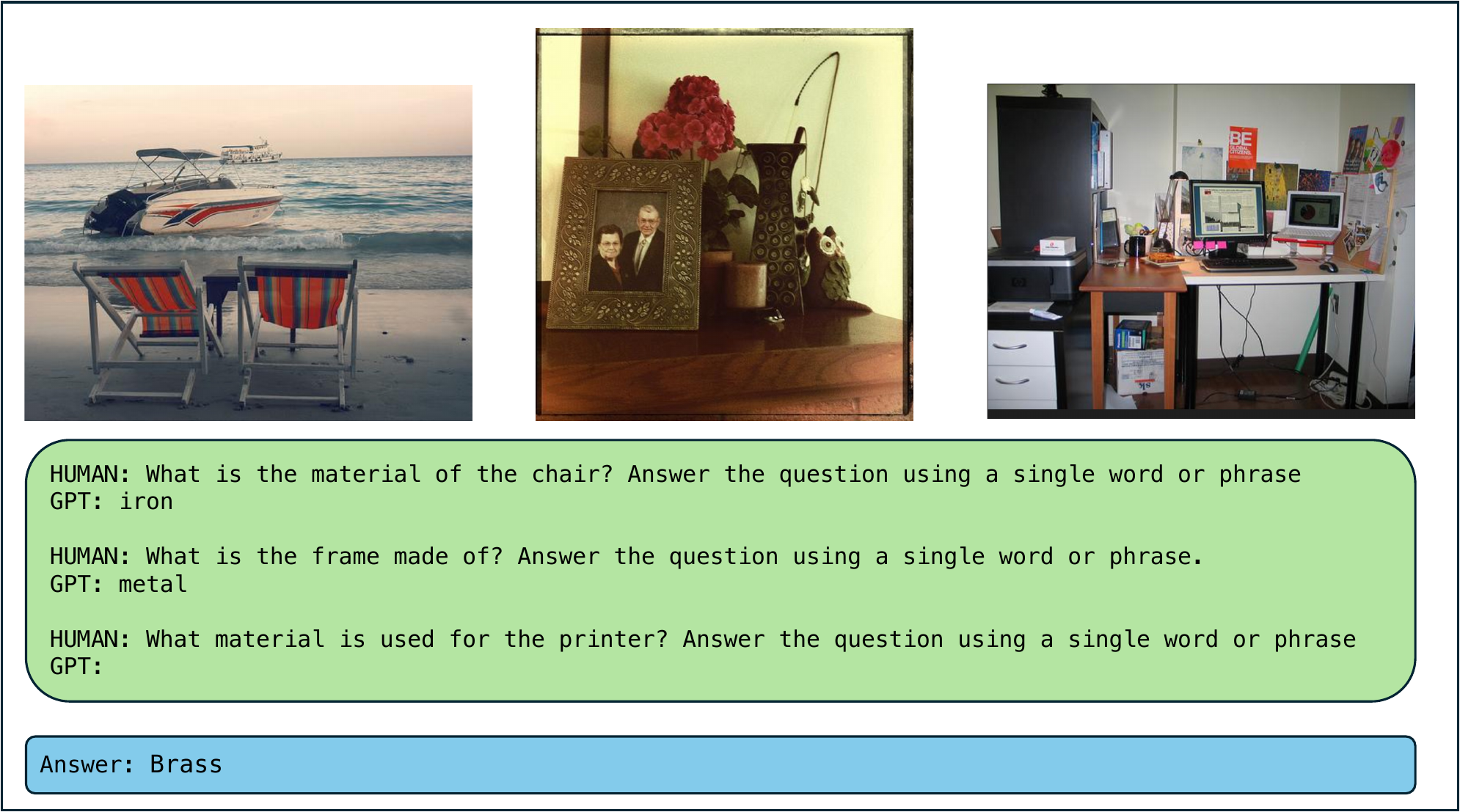}
        \caption{VL Checklist Question Answering task 0: Material}
    \end{subfigure}
    
    \begin{subfigure}{\textwidth}
        \centering
        \includegraphics[width=0.8\linewidth]{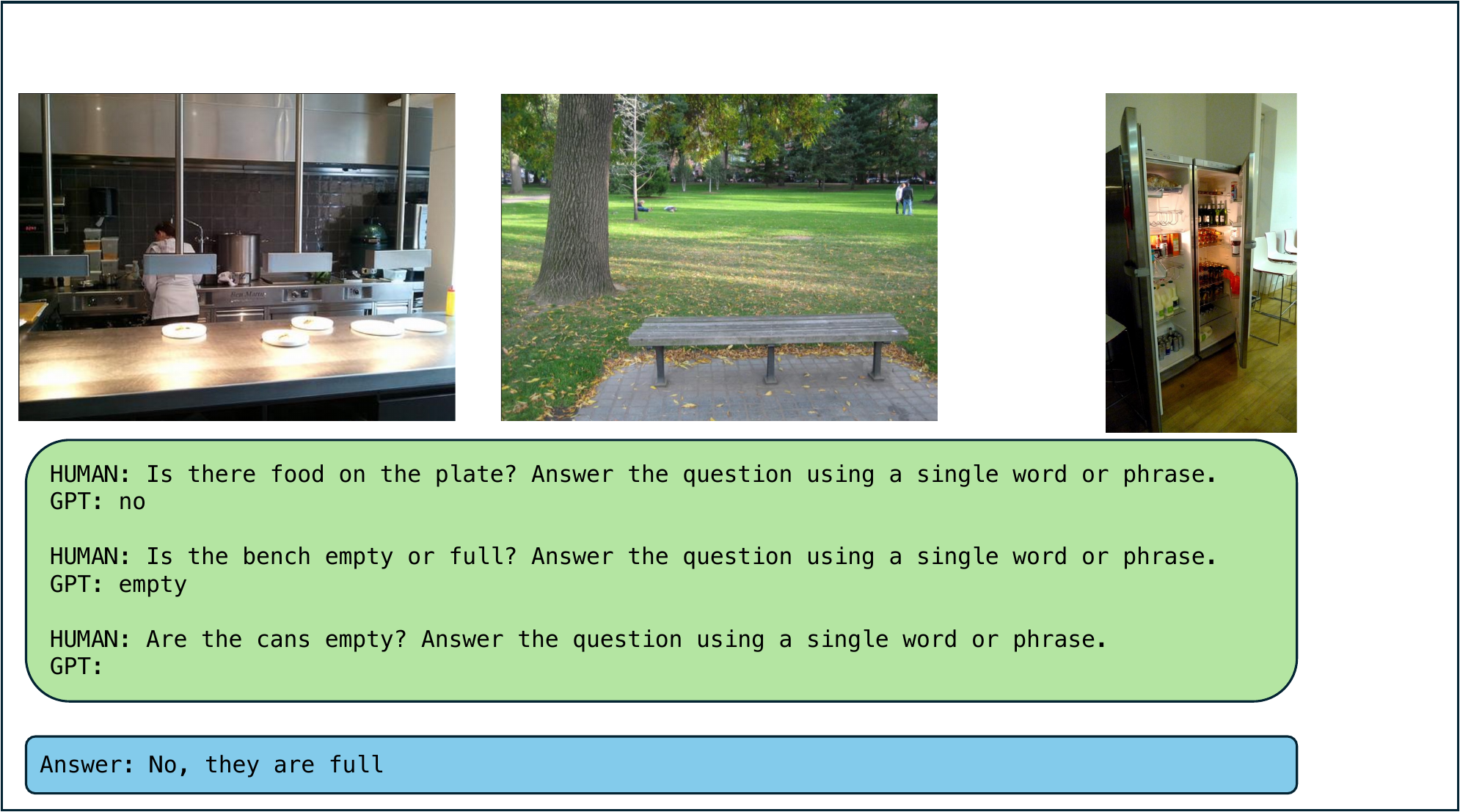}
        \caption{VL Checklist Question Answering task 4: State}
    \end{subfigure}
    \caption{VL Checklist Question Answering training examples tasks 0,4}
        \label{fig:vlqa04}
\end{figure}

\end{document}